\documentclass[sigconf]{acmart}
\usepackage{hyperref}
\makeatletter
\renewcommand\footnotetextcopyrightpermission[1]{}
\makeatother
\AtBeginDocument{%
  }

 \setcopyright{none}
\begin{document}

\title{\textbf{Foundation Models for Time Series: A Survey}}
\pagestyle{fancy}
\fancyfoot[C]{\thepage}  % Adds page number in the center of the footer

% \usepackage{amsmath,amssymb}
\author{Siva Rama Krishna Kottapalli\textsuperscript{\textsection}}
\email{siva.kottapalli@dell.com}
\affiliation{
  \institution{Dell Technologies}
  \city{Hopkinton}
  \state{MA}
  \country{USA}}

\author{Karthik Hubli\textsuperscript{\textsection}}
\email{karthik.hubli@dell.com}
\affiliation{
  \institution{Dell Technologies}
  \city{Hopkinton}
  \state{MA}
  \country{USA}}

\author{Sandeep Chandrashekhara\textsuperscript{\textsection}}
\email{sandeep.chandrashekara@dell.com}
\affiliation{
  \institution{Dell Technologies}
  \city{Hopkinton}
  \state{MA}
  \country{USA}}

\author{Garima Jain\textsuperscript{\textsection}}
\email{garima_jain@student.uml.edu}
\affiliation{
  \institution{University of Massachusetts Lowell}
  \city{Lowell}
  \state{MA}
  \country{USA}}

\author{Sunayana Hubli\textsuperscript{\textsection}}
\email{sunayana_hubli@student.uml.dell}
\affiliation{
  \institution{University of Massachusetts Lowell}
  \city{Lowell}
  \state{MA}
  \country{USA}}

\author{Gayathri Botla\textsuperscript{\textsection}}
\email{gayathri_botla@student.uml.edu}
\affiliation{
  \institution{University of Massachusetts Lowell}
  \city{Lowell}
  \state{MA}
  \country{USA}}

\author{Ramesh Doddaiah \textsuperscript{\textsection}}
\email{rdoddaiah@wpi.edu}
\affiliation{
  \institution{Worcester Polytechnic Institute
}
  \city{Worcester}
  \state{MA}
  \country{USA}}

\renewcommand{\shortauthors}{Kottapalli et al.}

\begin{abstract}
 Transformer-based foundation models have emerged as a dominant paradigm in time series analysis, offering unprecedented capabilities in tasks such as forecasting, anomaly detection, classification, trend analysis and many more time series analytical tasks. This survey provides a comprehensive overview of the current state of the art pre-trained foundation models, introducing a novel taxonomy to categorize them across several dimensions. Specifically, we classify models by their architecture design, distinguishing between those leveraging patch-based representations and those operating directly on raw sequences. The taxonomy further includes whether the models provide probabilistic or deterministic predictions, and whether they are designed to work with univariate time series or can handle multivariate time series out of the box. Additionally, the taxonomy encompasses model scale and complexity, highlighting differences between lightweight architectures and large-scale foundation models. A unique aspect of this survey is its categorization by the type of objective function employed during training phase. By synthesizing these perspectives, this survey serves as a resource for researchers and practitioners, providing insights into current trends and identifying promising directions for future research in transformer-based time series modeling.
\end{abstract}


\maketitle

% Move the footnote to the body of the document after \maketitle
\begingroup
\renewcommand\thefootnote{\textsection}
\endgroup

\pagestyle{plain}

\section{Introduction}
\footnotetext{All authors contributed equally to this research.}

Time series data is a key component of modern data analysis \cite{Shumway2000} and is commonly encountered across diverse range of fields including finance \cite{Tsay2005}, \cite{Tsay2014}, healthcare \cite{Penfold2013}, economics \cite{Nerlove2014}, climate science, inventory management \cite{Aviv2003}, \cite{Li2022}, energy management \cite{Deb2017}, traffic management \cite{Li2015}, internet of things (IOT) \cite{Cook2019}, industrial processes \cite{Cook2019}, supply chain optimization \cite{Aviv2003}, telecommunications, retail analytics \cite{Nunnari2017}, social media monitoring \cite{McCleary1980}, sensor networks \cite{Cook2019}, weather forecasting \cite{Campbell2005}, an even healthcare diagnostics \cite{Chui2017}. The importance of time series analysis lies in its ability to capture temporal dependencies and trends \cite{Eldele2021}, which is crucial for tasks ranging from imputation and classification to forecasting and anomaly detection. For instance, in finance, time series data can be used to predict stock prices or identify market anomalies \cite{Fama1970}, \cite{Stevenson2007}, while in healthcare, time series analysis can be applied to monitor patient vitals for early detection of medical conditions \cite{Penfold2013}, \cite{Chui2017} and forecast disease outbreaks. In meteorology \cite{Maturilli2013}, it is crucial for predicting weather patterns and climate changes, while in economics, time series analysis helps in forecasting key indicators like inflation and Gross Domestic Product (GDP) growth \cite{Nguyen2023}, \cite{Schneider1974}. In the energy sector, time series data is essential for optimizing demand forecasting \cite{Deb2017}, \cite{Lago2021} and resource allocation. Similarly, in inventory management, it aids in predicting stock levels \cite{Basu1996} and improving supply chain efficiency \cite{Li2022}. In telecommunications, time series data is used for network traffic analysis \cite{Bose2017} and predictive maintenance, while in retail, it supports demand forecasting \cite{Syntetos2005}, \cite{Morariu2018} and customer behavior analysis. In manufacturing, time series analysis can optimize production schedules \cite{Anderson2011} and monitor machine health, and in transportation, it helps predict traffic patterns \cite{Li2015} and optimize fleet management. Different applications of time series tasks including forecasting, clustering, imputation and more with and without fine-tuning a foundational model across different domains is shown in Figure 1.

\begin{figure*}[t]  % Use figure* to span both columns
  \centering
  \includegraphics[width=\textwidth]{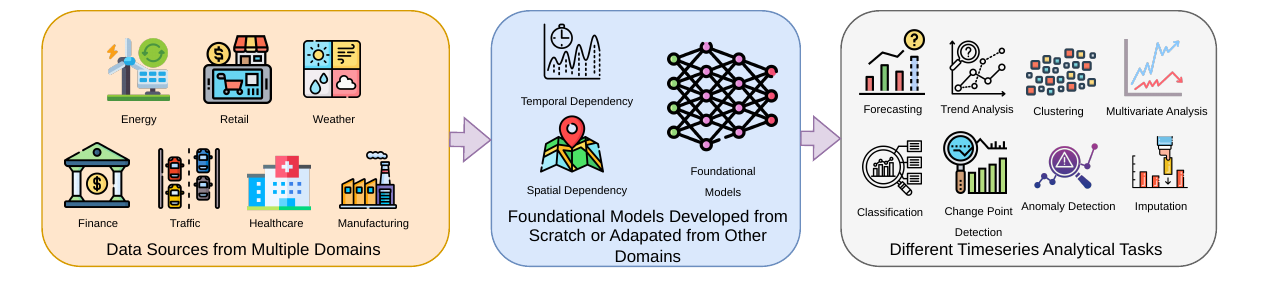}
  \caption{Training Foundation Time Series Models on Diverse Data Sources: Healthcare, Manufacturing, Finance and More.}
  \Description{Training Foundation Time Series Models on Diverse Data Sources: Healthcare, Manufacturing, Finance and More.}
  \label{fig:Fig1_TrainingFoundation}
\end{figure*}

Traditional approaches to time series analysis have largely relied on statistical methods \cite{Anderson2011}, \cite{Fuller2009}, such as Moving Averages (MA) \cite{moving_average_models}, which smooth data to identify underlying trends, and exponential smoothing, which gives more weight to recent observations. Methods like AutoRegressive Integrated Moving Average (ARIMA) models \cite{Stevenson2007}, \cite{Christodoulos2010}, which combine AutoRegressive (AR) and Moving Average (MA) components \cite{moving_average_models}, are commonly used to model time-dependent structures. Seasonal and Trend decomposition using Loess (STL) \cite{Cleveland1990}, which separates a time series into trend, seasonal, and residual components, is also a widely used technique in time series analysis. In Figure 2, we see the intraday price movements of Apple Inc. (AAPL) using a 5-minute time frame. The analysis incorporates two types of moving averages: the 15-period Simple Moving Average (SMA) and the 15-period Exponential Moving Average (EMA). These two indicators are commonly used in technical analysis to smooth out price data and identify potential trends.

While these traditional methods are effective in many cases, they often struggle with more complex, non-linear patterns or high-dimensional data. In contrast, machine learning algorithms like Support Vector Machines (SVMs) \cite{Hearst1998} and Gradient Boosting Machines (GBMs) \cite{Friedman2001} have proven effective at capturing more intricate temporal dependencies, particularly when the data exhibits linear or relatively simple trends. However, these methods can face significant challenges when dealing with irregularly sampled data or when manual feature engineering is needed to extract useful temporal features. For example, ARIMA models may have difficulty capturing long-range dependencies or non-linear relationships in the data \cite{Kolambe2024}, while traditional machine learning algorithms often require careful preprocessing and feature selection to be effective in time series tasks \cite{Bouktif2018}.

\begin{figure}[t]  % Use figure for a single column
  \centering
  \includegraphics[width=\columnwidth]{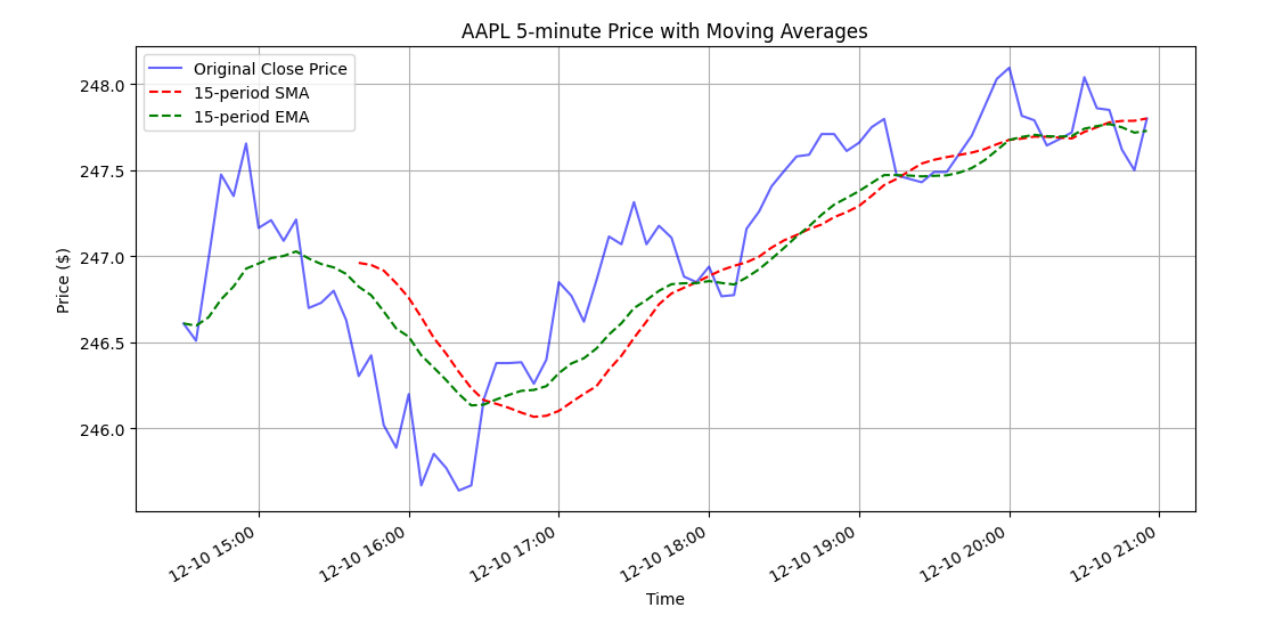}  % Adjust width to column width
  \caption{Intraday 5-Minute Price Movements of Apple (AAPL) with 15-Period Simple Moving Average (SMA) and Exponential Moving Average (EMA)}
  \Description{Figure 2: Intraday 5-Minute Price Movements of Apple (AAPL) with 15-Period Simple Moving Average (SMA) and Exponential Moving Average (EMA)}
  \label{fig:Fig2_AAPL}
\end{figure}

Furthermore, these traditional models typically operate under certain assumptions, such as the data being stationary or evenly spaced, which can be restrictive in many practical scenarios \cite{Shumway2000}. When the temporal structure is more intricate, or the data points are not uniformly spaced (as in the case of sensor data or stock market data with irregular trading hours), the traditional models fail to deliver optimal results \cite{Atluri2018}. This limitation has driven the need for more advanced techniques that can handle these complexities effectively.

\subsection{Neural Networks for Time Series Analysis}

In response to these limitations, neural networks, particularly Recurrent Neural Networks (RNNs) \cite{Connor1994}, \cite{Hewamalage2021} and Convolutional Neural Networks (CNNs) \cite{Borovykh2017}, \cite{Zheng2014} have emerged as powerful alternatives for time series modeling. These neural networks are particularly appealing because they have the ability to learn hierarchical representations of data directly from raw inputs, eliminating the need for extensive feature engineering \cite{Najafabadi2015}. This “end-to-end” learning approach enables neural networks to automatically capture the underlying structure of the data, which is especially valuable in time series forecasting, anomaly detection, and classification, where temporal dependencies often exhibit complex, non-linear patterns \cite{Usmani2024}. 
RNNs, introduced by Rumelhart et al. in the 1980s, are designed to process sequential data by maintaining hidden states that capture information about previous inputs in the sequence \cite{Rumelhart1986}. At each time step, the RNN updates its hidden state based on both the current input and the previous state, allowing it to model temporal relationships in the data \cite{Esteban2016}. This makes RNNs a natural fit for time series tasks where the prediction at any time step depends on earlier observations. For instance, in applications such as stock price prediction, weather forecasting, or sensor data analysis, the value of the time series at a given point is influenced by previous time points, and RNNs are well-suited to capture this dynamic \cite{Mienye2024}.
However, despite their theoretical advantages, traditional RNNs face several critical limitations, most notably the vanishing gradient problem. When training RNNs via Back Propagation Through Time (BPTT), gradients can become exceedingly small as they are propagated backward through long sequences \cite{Kag2021}. This causes the model to struggle with learning long-term dependencies, especially in sequences where the relationship between events is separated by many time steps. As a result, traditional RNNs tend to perform poorly when applied to tasks that require understanding of long-range dependencies, such as forecasting long-term trends or detecting anomalies in large time series datasets \cite{Choi2021}.
To address the shortcomings of basic RNNs, Long Short-Term Memory (LSTM) networks were introduced by Hochreiter and Schmidhuber in 1997. LSTMs are specifically designed to mitigate the vanishing gradient problem by incorporating memory cells and gating mechanisms that control the flow of information. Unlike traditional RNNs, LSTMs use three main gates—the input gate, output gate, and forget gate—to selectively remember or forget information at each time step. This allows LSTMs to capture long-term dependencies more effectively, making them highly successful in applications such as speech recognition, machine translation, and time series forecasting \cite{Hochreiter1997}.
Similarly, Gated Recurrent Units (GRUs), introduced by Cho et al. in 2014, are a simplified variant of LSTMs that use two gates (reset and update) to manage the flow of information. While LSTMs have been widely adopted due to their ability to capture complex temporal dependencies, GRUs have gained popularity because they are computationally more efficient and require fewer parameters \cite{Chung2014}. Both LSTMs and GRUs have been proven to outperform traditional RNNs in many time series applications, such as predicting stock prices, energy demand forecasting, and anomaly detection in sensor networks.
Despite their successes, RNN-based architectures, including LSTMs and GRUs, still face several significant challenges that limit their scalability and efficiency in handling large-scale time series data:

\begin{enumerate}
    \item \textbf{Sequential Nature and Parallelization}: RNNs process input data one time step at a time, which makes them inherently difficult to parallelize during training \cite{Schmidt2019}. Unlike feedforward neural networks, where all operations can be performed in parallel, RNNs rely on the computation of one step’s output before moving to the next, making training and inference slower, especially for long sequences. This sequential processing leads to high computational costs and longer training times, particularly when working with large datasets or complex models.
    
    \item \textbf{Long-Term Dependencies}: Although LSTMs \cite{Hochreiter1997} and GRUs \cite{Chung2014} are designed to capture long-term dependencies, they can still struggle with extremely long sequences or sequences with highly irregular patterns. In some cases, the gradients may still vanish or explode during backpropagation, making it difficult for the models to retain important information over many time steps \cite{Hochreiter2001}. While LSTMs and GRUs mitigate the vanishing gradient problem compared to traditional RNNs, they are not immune to it, especially in cases where the input sequence is exceedingly long or contains complex patterns.

    \item \textbf{Memory and Computation Constraints}: Training RNNs and its variants LSTMs, and GRUs on large-scale datasets requires significant memory and computational resources \cite{Yang2020}. Due to their sequential processing, these models may require extensive amounts of time and computational power to process long sequences or large batches of data. In real-time or resource-constrained environments, this can become a significant bottleneck, limiting the practical deployment of these models in many time-sensitive applications \cite{Najafabadi2015}.
    
    \item \textbf{Overfitting and Generalization}: RNN-based models can also suffer from overfitting when trained on small datasets, particularly when the network architecture is too complex relative to the amount of data available \cite{Schmidt2019}. The large number of parameters in LSTMs and GRUs increases the risk of overfitting, which can lead to poor generalization performance when applied to unseen data \cite{Najafabadi2015}.
\end{enumerate}

\subsection{The Transformer Paradigm}

The introduction of Transformers \cite{Vaswani2017} in 2017 represents a major paradigm shift in sequence modeling. Transformers, initially developed for Natural Language Processing (NLP) tasks, rely on a novel self-attention mechanism to capture dependencies between elements in a sequence, without the need for recurrence. Unlike RNNs, Transformers process all elements in the sequence simultaneously, allowing for more efficient parallelization and faster training times \cite{Vaswani2017}. This self-attention mechanism enables the model to focus on different parts of the input sequence dynamically, which is particularly useful when capturing long-range dependencies that might span large portions of the data.

The Transformer architecture’s ability to model long-range dependencies with a reduced computational cost is a significant advantage over traditional RNNs. This ability has been crucial in tackling the challenges inherent in time series data, such as irregular sampling intervals or highly complex, non-linear patterns that span across different time scales. Additionally, Transformers do not suffer from the vanishing gradient problem to the same extent as RNNs, because they do not rely on the sequential processing of inputs. Instead, they allow for direct connections between any two elements in the sequence through the self-attention mechanism, facilitating the learning of more complex temporal relationships.

As a result, Transformer-based models have quickly gained traction in the field of time series analysis \cite{Wen2023}, outperforming traditional methods and even RNN-based architectures in many tasks, from forecasting to anomaly detection. In recent years, a number of Transformer variants, such as the \textit{Time Series Transformer (TST)} \cite{Jin2022} and \textit{Informer} \cite{Zhou2021}, have been specifically designed to address the unique challenges posed by time series data, such as the need to handle long sequences efficiently and deal with irregularly spaced data points.

\subsection{Transformers: Foundation Models for Time Series}

Traditional methods, such as training a neural network on each individual time series, often fail to yield good results, particularly when dealing with a large number of time series or varying temporal patterns \cite{Zeng2023}. This is because training models independently for each time series ignores the potential for shared temporal patterns and commonalities across different series \cite{Lai2018}. In contrast, transformer-based models, which are designed to learn from the entire dataset simultaneously, can capture these shared patterns and provide a more robust and generalized solution. By leveraging a foundational model that can learn representations across multiple time series, Transformers are better equipped to make more accurate predictions, even when faced with diverse and complex data \cite{Zerveas2021}.

With these advantages in mind, there is a growing interest in exploring Transformer-based architectures for time series analysis. By leveraging these models, researchers and practitioners aim to develop more accurate, efficient, and scalable solutions for time series forecasting, anomaly detection, classification, and beyond. The success of Transformers in other domains like NLP \cite{Touvron2023}, \cite{OpenAI2023} and computer vision \cite{Khan2021} further reinforces the potential of this architecture for revolutionizing time series analysis, offering a compelling alternative to traditional methods and RNN-based architectures.

The rapid advancements in Transformer-based models and their continued success in various domains suggest that they are not just a passing trend but a fundamental shift in how sequence data, including time series, can be modeled \cite{Karita2019}. Consequently, Transformer-based architectures are poised to play a pivotal role in shaping the future of time series analysis, offering new avenues for improving forecasting accuracy, anomaly detection capabilities, and the general understanding of temporal data across diverse fields \cite{Atluri2018}, \cite{Wen2023}, \cite{Ma2023}.

\section{Background}
\label{background}

\subsection{Unique Characteristics of Time Series Data}

Time series data exhibits unique characteristics that distinguish it from other forms of data, presenting both challenges and opportunities for analysis and modeling \cite{Shumway2000}. This section explores these characteristics and their implications, along with a discussion of the diverse real-world applications where time series data plays a critical role.

\begin{figure}[t]  % Use figure for a single column
  \centering
  \includegraphics[width=\columnwidth]{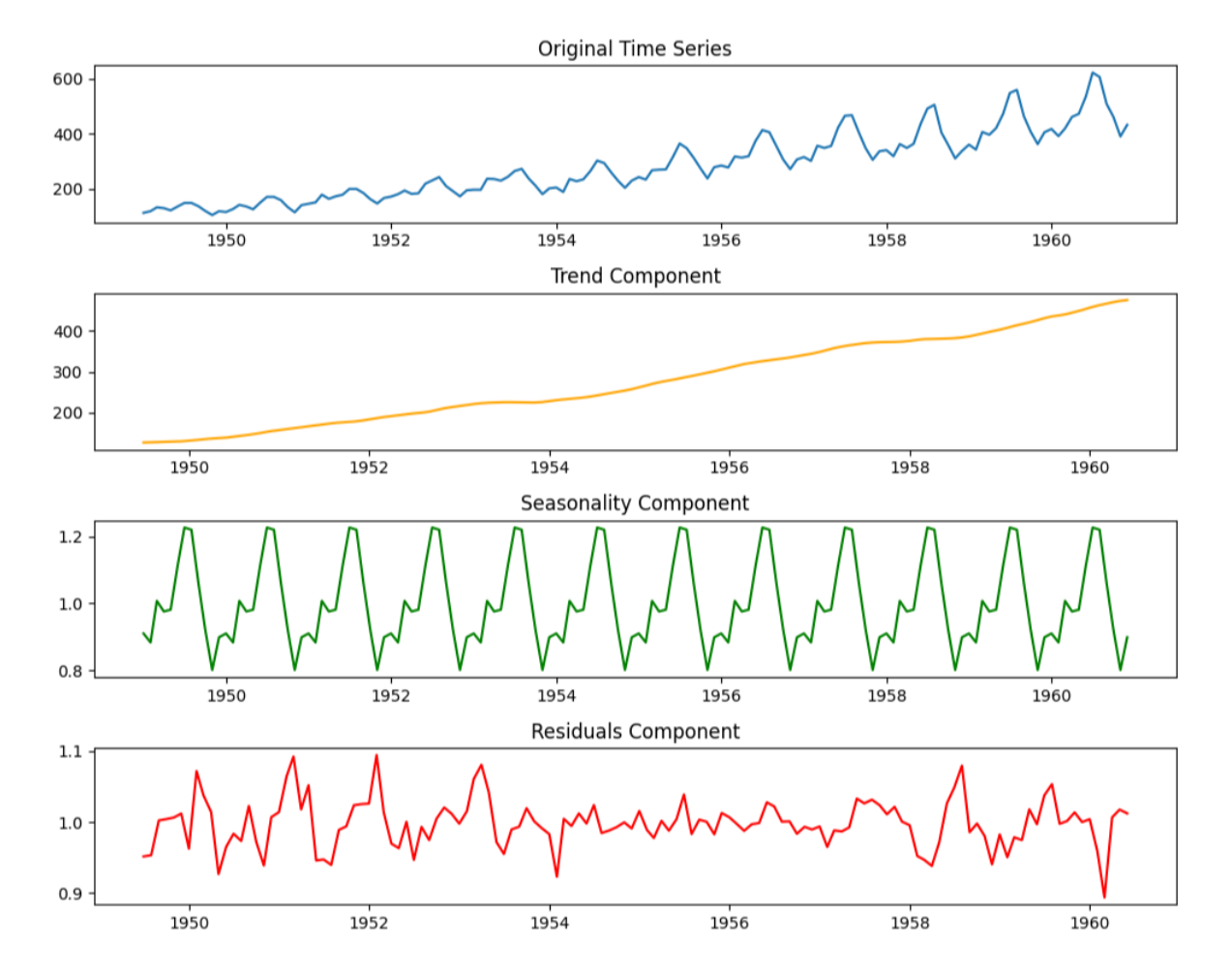}  % Adjust width to column width
  \caption{Seasonal Decomposition of Airline Passenger Data: Trend, Seasonality, and Residuals. Trend reflects the long-term direction in the data. Seasonality captures repeating patterns at regular intervals. Residuals represent the random fluctuations left after removing the trend and seasonality}
  \Description{Figure 3: Seasonal Decomposition of Airline Passenger Data: Trend, Seasonality, and Residuals. Trend reflects the long-term direction in the data. Seasonality captures repeating patterns at regular intervals. Residuals represent the random fluctuations left after removing the trend and seasonality}
  \label{fig:Fig3_Seasonal}
\end{figure}

\subsubsection{Sequential Nature}

Time series data is inherently sequential, where observations are recorded over time in a specific order \cite{Box1976}. This sequential nature implies that each data point is not independent but is influenced by its temporal context. Unlike static datasets, the sequence of observations conveys meaningful information that must be preserved during analysis. For example, the daily stock price of a company is influenced by prior trends, patterns, and market behaviors. Modeling techniques must account for this order to capture dependencies and make accurate predictions \cite{Fama1970}.

\subsubsection{Temporal Dependencies}

A defining feature of time series data is the presence of temporal dependencies \cite{Lai2018}. These dependencies manifest in two forms:

\begin{itemize}
    \item \textbf{Short-term dependencies}: The immediate past significantly influences future observations. For instance, today’s weather conditions impact tomorrow’s forecast.
    \item \textbf{Long-term dependencies}: Distant past observations can also affect the current state, as seen in seasonal patterns in sales data or recurring events in climate studies. Modeling these dependencies requires techniques that can identify and quantify relationships across varying time scales, which is a critical challenge in time series analysis.
\end{itemize}
To better understand the dynamics of these temporal dependencies, the Figure \textit{3} illustrates how both short-term and long-term patterns emerge from airline passenger time series dataset. The plots demonstrate the underlying trend, seasonal cycles, and irregular fluctuations, helping to visualize both immediate and long-range influences on the data. The trend component shows the long-term movement of the data, the seasonal component shows recurring patterns in the short term, and the residual component shows random and unpredictable noise that is harder to capture by either trend or seasonal components.

\subsubsection{Multivariate Complexities}

Many real-world time series datasets involve multiple variables recorded simultaneously, leading to multivariate time series \cite{Tsay2014}. These variables often interact and influence one another. For example, in healthcare, the interaction between a patient’s heart rate, blood pressure, and oxygen levels forms a multivariate time series \cite{Pope1999}. Effectively analyzing such datasets requires models capable of capturing both intra-variable temporal dependencies and inter-variable interactions, adding another layer of complexity.

\subsubsection{Irregular Sampling and Missing Data}

Time series data is not always recorded at regular intervals, resulting in irregular sampling \cite{Rehfeld2011}. This irregularity can occur due to equipment malfunctions, human errors, or external disruptions. Furthermore, missing data is a frequent issue, particularly in real-world settings. For instance, Internet of Things (IoT) sensors may fail to record data due to connectivity issues, leading to gaps in the time series \cite{Cook2019}. Handling such challenges often involves imputation techniques or model adaptations to account for irregularities and incomplete data.

\subsubsection{Noise and Non-Stationarity}

Time series data is frequently noisy, with random fluctuations that obscure underlying patterns \cite{Shumway2000}. Additionally, many time series datasets are non-stationary, meaning their statistical properties, such as mean and variance, change over time. For example, economic data may shift due to inflation rates or policy changes, and seasonal sales patterns may evolve with changing consumer behavior \cite{Nerlove2014}. Addressing noise and non-stationarity requires robust preprocessing techniques and adaptive models.

\subsubsection{High Dimensionality in Long Sequences }

Time series datasets can grow significantly in size when spanning long timeframes or involving high-frequency sampling \cite{Gu2021}. For example, minute-by-minute trading data collected over several years can result in millions of data points, creating computational and storage challenges \cite{Schadt2010}. Efficient handling of such datasets necessitates scalable modeling approaches like LSTM, 1D-CNN and Transformers that can process long sequences without losing critical information.

\subsection{Key Innovations of Transformers}

The advent of Transformer \cite{Vaswani2017} models has revolutionized the field of machine learning, particularly in domains involving sequential data, such as natural language processing (NLP) \cite{Zhao2023}. Unlike traditional architectures, such as recurrent neural networks (RNNs) \cite{Rumelhart1986} and long short-term memory networks (LSTMs) \cite{Hochreiter1997}, Transformers \cite{Vaswani2017} introduce several key innovations that address critical limitations of previous approaches. This section highlights the two most significant innovations: the attention mechanism and the scalability advantages, which together make transformers a compelling choice for time series modeling.

\subsubsection{Attention Mechanism and its Role in Sequential Data}

The attention mechanism lies at the heart of Transformer architecture \cite{Vaswani2017} shown in Figure \textit{4}, enabling models to process and extract relationships between data points without relying on strict sequential computations. Traditional models like RNNs and LSTMs suffer from vanishing gradients and limited ability to capture long-range dependencies, as they inherently process data sequentially \cite{Salehinejad2017}. The attention mechanisms shown in Figure \textit{5} overcomes these challenges by offering a global perspective of the input sequence. In the context of time series, the attention mechanisms provide the following advantages \cite{Wen2023}, \cite{Zeng2023}, \cite{Lin2021}:

\begin{itemize}
    \item \textbf{Long-Range Dependency Modeling}: Time series data often exhibits long-range temporal dependencies, such as seasonal patterns or recurring events \cite{Wen2023}. The self-attention mechanism in Transformers computes pairwise relationships between the time steps, allowing the model to focus on relevant inputs from both distant and recent time points \cite{Vaswani2017}.
    \item \textbf{Dynamic Weighting}: Instead of assigning fixed weights based on positional order, attention dynamically learns weights that emphasize the most relevant time steps \cite{Lin2021}. This ensures that the model prioritizes meaningful observations, such as anomalies or sudden changes in trends, which are critical in applications like anomaly detection in industrial sensors \cite{Lin2021}.
    \item \textbf{Context-Aware Representations}: By jointly attending to all time steps, Transformers create context-aware embeddings, which integrate information from the entire sequence. These embeddings are particularly useful in multivariate time series, where the interactions between variables are crucial for prediction and analysis \cite{Zerveas2021}.
\end{itemize}
 
Mathematically, the self-attention mechanism computes the attention score as follows \cite{Vaswani2017}:

\begin{figure}[t]  % Use figure for a single column
  \centering
  \includegraphics[width=\columnwidth]{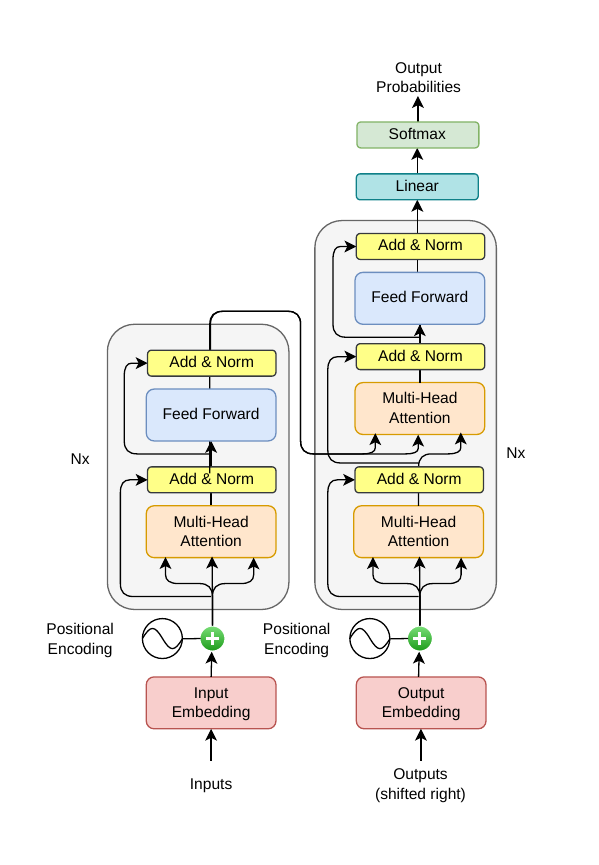}  % Adjust width to column width
  \caption{Transformer – model architecture from “Attention Is All You Need” paper \cite{Vaswani2017}}
  \Description{Figure 4: Transformer – model architecture from “Attention Is All You Need” paper \cite{Vaswani2017}}
  \label{fig:Fig4_Transformer}
\end{figure}

$$Attention(Q,\ K,\ V) = softmax\left( \frac{QK^{T}}{\sqrt{d_{k}}} \right)V$$

where Q, K, and V represent Query, Key, and Value matrices derived from the input sequence, and dk is the dimensionality of the key vectors. This formulation enables transformers to efficiently compute relationships across the entire sequence in parallel, a stark contrast to the step-by-step operations in RNNs \cite{Karita2019}.

\begin{figure}[t]  % Use figure for a single column
  \centering
  \includegraphics[width=\columnwidth]{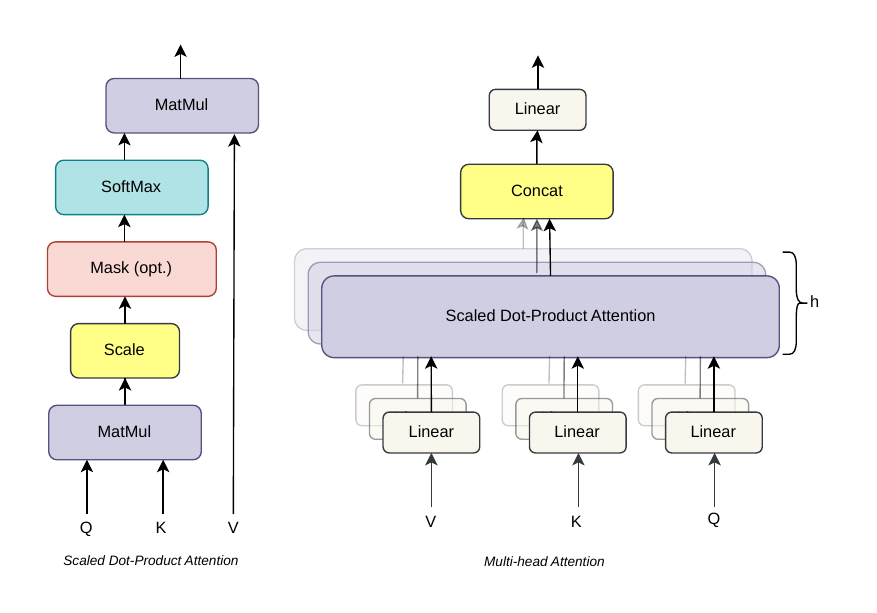}  % Adjust width to column width
  \caption{Transformer – Scaled Dot-Product Attention (left) and Multi-Head Attention (right) from “Attention Is All You Need” paper \cite{Vaswani2017}}
  \Description{Figure 5: Transformer – Scaled Dot-Product Attention (left) and Multi-Head Attention (right) from “Attention Is All You Need” paper \cite{Vaswani2017}}
  \label{fig:Fig5}
\end{figure}

\subsubsection{Scalability and Parallelism Advantages Over Traditional Architectures}

One of the most transformative aspects of the Transformer model is its ability to leverage parallelism, making it significantly more scalable compared to traditional sequential architectures \cite{Sanford2024}. This scalability stems from two critical factors:

\begin{itemize}
    \item \textbf{Non-Sequential Processing}: Unlike RNNs, which process inputs one time step at a time, Transformers operate on the entire sequence simultaneously \cite{Vaswani2017}. This non-sequential processing eliminates the dependency on previous computations, allowing for efficient utilization of modern hardware accelerators like GPUs and TPUs.
    \item \textbf{Efficient Handling of Long Sequences}: Time series datasets often involve long sequences, posing challenges for RNNs \cite{Rumelhart1986} and LSTMs \cite{Hochreiter1997} due to their inherent limitations in memory and computation. Transformers, on the other hand, process all time steps in parallel, ensuring consistent computational efficiency regardless of sequence length \cite{Sanford2024}. This advantage is especially relevant in applications like high-frequency financial trading, where datasets consist of millions of observations.
\end{itemize}
However, the scalability of Transformers is not without challenges. The self-attention mechanism requires O(n2) memory and computation for a sequence of length n, which can become prohibitive for extremely long time series \cite{Wu2021}. To address this, recent innovations such as sparse attention and linear Transformers have been proposed, reducing the complexity to O(n) \cite{Rabe2021}, \cite{Wang2020} in certain cases. These advancements extend the applicability of Transformers to resource-constrained environments.

\subsubsection{Implications for Time Series Modeling}

The innovations introduced by Transformers have significant implications for time series analysis \cite{Ahmed2023}, \cite{Wen2023}. The attention mechanism enables models to capture complex temporal dynamics, such as seasonality and long-term dependencies, while the scalability ensures that these models remain practical for large-scale datasets. Furthermore, the ability to process multivariate inputs and derive context-aware representations positions Transformers as a versatile tool for tackling a wide range of time series tasks, from forecasting and classification to anomaly detection \cite{Zerveas2021}. The introduction of the attention mechanism and the emphasis on scalability mark a paradigm shift in the handling of sequential data. These innovations not only address the limitations of traditional architectures but also pave the way for the development of more robust and efficient models tailored to the unique characteristics of time series data.

\subsection{Time Series Applications}

Time series data has become ubiquitous in modern analytics, encompassing domains like finance, healthcare, manufacturing, and climate science. Transformer-based foundation models have demonstrated remarkable potential in addressing a variety of time series tasks. This section delves into key applications of time series analysis, including forecasting, imputation, anomaly detection, classification, change point detection, and clustering, highlighting how Transformers are revolutionizing these areas \cite{Shumway2000}.

\subsubsection{Time Series Forecasting}

Forecasting is one of the most critical applications of time series analysis, involving the prediction of future values based on historical data \cite{Newbold1974}. Accurate forecasting enables informed decision-making across industries such as energy demand planning, stock market predictions, storage capacity forecasting, and weather forecasting. Transformer-based models have significantly advanced time series forecasting by overcoming the limitations of traditional statistical methods and recurrent models \cite{Wen2023}. Their attention mechanism allows for learning long-term dependencies, capturing complex temporal dynamics across varying scales. Models like \textit{Informer} \cite{Zhou2021} and \textit{Autoformer} \cite{Wu2021} optimize for scalability and computational efficiency, making them highly suitable for handling high-dimensional time series data with irregular patterns. In Figure 6 forecasting is performed using SARIMA (Seasonal AutoRegressive Integrated Moving Average), a traditional time series modeling technique. 

\begin{figure}[t]  % Use figure for a single column
  \centering
  \includegraphics[width=\columnwidth]{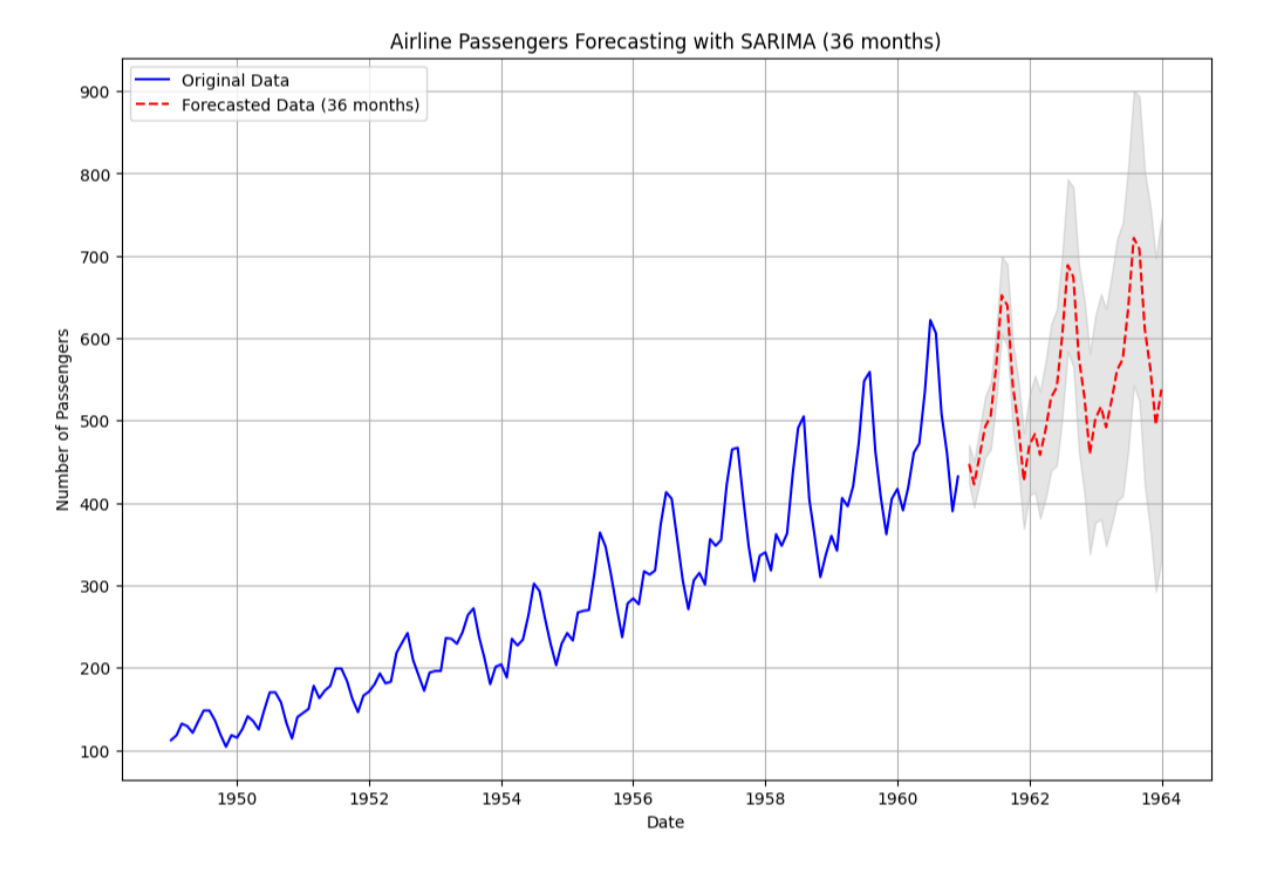}  % Adjust width to column width
  \caption{Forecasting Airline Passenger Numbers for the Next 36 Months using SARIMA Model}
  \Description{Forecasting Airline Passenger Numbers for the Next 36 Months using SARIMA Model}
  \label{fig:Fig6}
\end{figure}

\subsubsection{Time Series Imputation}

Imputation addresses the challenge of missing or corrupted data, a common issue in real-world time series \cite{Adhikari2022}, \cite{Fang2020}. Filling these gaps accurately is vital for downstream applications, as erroneous imputations can propagate errors through analytical pipelines. Transformers, with their inherent ability to model complex dependencies, excel in learning contextual relationships to infer missing values. Techniques such as bidirectional attention \cite{Ni2022}, \cite{Yang2021} and encoder-decoder frameworks \cite{Wang2024}, \cite{Liu2024} have been employed to reconstruct missing segments effectively. For instance, models like \textit{TimeTransformer} \cite{Liu2024} utilize self-attention mechanisms to predict missing data points in multidimensional datasets.

\subsubsection{Anomaly Detection}

Anomaly detection focuses on identifying deviations from normal patterns. Anomaly detection is crucial for not only time series applications such performance impact detection \cite{Pope1999}, but also NLP applications like fraud detection, performance impact, equipment fault diagnosis and cybersecurity \cite{Cook2019}, \cite{Schmidl2022}, \cite{Xu2021}. The ability to detect rare or subtle anomalies amidst vast amounts of time series data is a key challenge. In Figure \textit{7} anomaly detection task is performed on time series data. Transformer models provide a powerful framework for anomaly detection due to their capacity for learning contextual representations \cite{Xu2021}, \cite{Chen2021}. Pretrained models can be fine-tuned for anomaly detection tasks by leveraging embeddings that capture normal behavior patterns. Advanced approaches, such as using Transformers in conjunction with variational autoencoders \cite{Kingma2019} or generative adversarial networks \cite{Goodfellow2020}, further enhance anomaly detection capabilities by enabling unsupervised or semi-supervised learning.

\begin{figure}[t]  % Use figure for a single column
 \centering
  \includegraphics[width=\columnwidth]{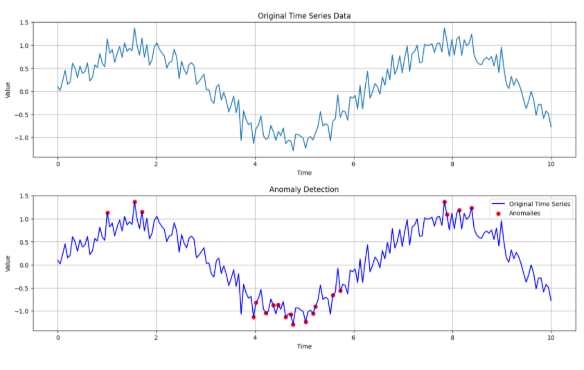}  % Adjust width to column width
  \caption{Anomaly Detection in Time Series Data: Identification of Anomalies Highlighted by Red Dots}
  \Description{Anomaly Detection in Time Series Data: Identification of Anomalies Highlighted by Red Dots}
  \label{fig:Fig7}
\end{figure}

\subsubsection{Time Series Classification}

Classification of time series data involves assigning labels to sequences, with applications ranging from activity recognition in wearable devices to disease diagnosis using medical signals \cite{Fawaz2019}, \cite{Fawaz2018}, \cite{Jin2023}. High inter-class similarity and intra-class variability pose challenges for effective classification. Transformer-based models address these challenges through their ability to capture high-resolution temporal and spatial features. By utilizing pretrained embeddings and self-attention mechanisms, these models can identify critical subsequences and patterns relevant to classification tasks \cite{Lu2022}. Approaches like hierarchical attention mechanisms have shown promise in improving accuracy for multi-scale time series. 

\subsubsection{Change Point Detection}

Change point detection identifies moments when the statistical properties of a time series shift, indicating regime changes or transitions \cite{VandenBurg2020}, \cite{Aminikhanghahi2017}. This application finds utility in detecting financial market shifts, climate pattern changes, and network traffic anomalies. Transformers enable more precise change point detection by leveraging attention mechanisms to model dependencies across long time horizons. They can learn intricate temporal relationships that often precede change points. Recent approaches integrate domain-specific knowledge with pretrained Transformer models to improve detection accuracy and interpretability.

\subsubsection{Time Series Clustering}

Clustering aims to group time series based on similarity, facilitating insights into underlying patterns. Applications span customer segmentation, biological signal analysis, and sensor network optimization \cite{McLachlan1988}, \cite{Liao2005}, \cite{Aghabozorgi2015}. Transformers enhance clustering through their ability to generate meaningful sequence embeddings that capture both temporal and feature-level similarities. Pretrained embeddings from models such as \textit{TS-TCC} (Time Series Transformer for Clustering and Classification) are particularly effective, offering robust representations that can be fed into standard clustering algorithms like k-means or hierarchical clustering \cite{Eldele2021}.

Transformer-based models have made significant contributions across various time series applications, each offering unique advantages as highlighted in the below Table \textit{1}.

 \begin{table*}[t]
\centering
\caption{Key Contributions of Transformer-based Models Across Time Series Applications}
\begin{tabular}{|p{3cm}|p{8cm}|}
\hline
\textbf{Application} & \textbf{Key Transformer Contribution} \\
\hline
Forecasting & Long-term dependency modeling, multivariate scaling \\
\hline
Imputation & Contextual learning for missing data reconstruction \\
\hline
Anomaly Detection & Contextual embedding-based outlier detection \\
\hline
Classification & Multiscale pattern recognition \\
\hline
Change Point Detection & Accurate shift modeling over long time horizons \\
\hline
Clustering & Representation learning for similarity-based tasks \\
\hline
\end{tabular}
\label{tab:transformer-contributions}
\end{table*}

\subsection{Foundation Models for Time Series}

Foundation models represent a transformative concept in machine learning, characterized by their ability to learn generalizable representations through pretraining on large-scale datasets and their subsequent fine-tuning for specific tasks. Originating from natural language processing (NLP) \cite{Wolf2020} with models like \textit{LLAMA \cite{Touvron2023}}, \textit{BERT \cite{Devlin2018}} and \textit{GPT \cite{OpenAI2023}}, the foundation model paradigm is now being extended to time series analysis. This section explores the defining characteristics of foundation models and highlights notable Transformer-based examples tailored for time series data.

\subsubsection{Characteristics of Foundation Models}

The defining trait of foundation models is their ability to act as a universal backbone for diverse downstream tasks. This capability is achieved through a two-stage process: pretraining and fine-tuning.

\begin{itemize}
    \item \textbf{Pretraining}: Foundation models are trained on vast datasets to capture generalized patterns, structures, and dependencies \cite{Bommasani2021}. In the context of time series, pretraining involves learning representations that encapsulate temporal dependencies, multivariate interactions, and periodic patterns across diverse datasets. These pretrained models are designed to understand common structures in time series data, such as seasonality or trend shifts, without being tied to a specific task \cite{Wen2023}.
    \item \textbf{Fine-Tuning Across Tasks}: After pretraining, the model can be fine-tuned on specific tasks such as forecasting, anomaly detection, or classification. Fine-tuning allows the model to adapt its generalized knowledge to the idiosyncrasies of a particular dataset or task, often with minimal labeled data \cite{Bommasani2021}.
\end{itemize}
The ability of foundation models to generalize stems from several key properties:

\begin{enumerate}
    \item \textbf{Task-Agnostic Pretraining Objectives}: Pretraining often employs self-supervised learning objectives, such as next-step prediction, masked reconstruction, or contrastive learning. These objectives enable the model to learn meaningful representations without requiring extensive labeled data, a critical advantage in time series domains where labeled data can be scarce \cite{Wen2023}.
    \item \textbf{Scalability Across Domains}: Foundation models are typically trained on heterogeneous datasets spanning multiple domains, such as healthcare, finance, and climate. This diversity enhances their robustness and transferability to unseen tasks \cite{Yuan2022}.
    \item \textbf{Adaptability Through Fine-Tuning}: Fine-tuning offers flexibility, allowing pretrained models to be specialized for domain-specific nuances or integrated with domain knowledge, such as physical constraints in climate modeling or business rules in financial forecasting \cite{Han2024}.
\end{enumerate}
 
\subsubsection{Notable Transformer-Based Foundation Models for Time Series}

Several Transformer-based models have emerged as foundation models for time series, each leveraging the Transformer’s strengths to tackle the unique challenges of time series data. Below, we discuss a few notable examples that illustrate the breadth and potential of this approach.

\textit{Time Series Transformer} (\textit{TST}) adapts the standard Transformer architecture for time series by employing positional encodings tailored to temporal data and attention mechanisms to capture long-term dependencies. Pretrained on diverse datasets, \textit{TST} can be fine-tuned for tasks ranging from anomaly detection in industrial sensors to forecasting in energy systems \cite{Wen2023}. \textit{Informer }\cite{Zhou2021} is specifically designed to address the scalability challenges posed by long time series. By introducing a sparse self-attention mechanism, \textit{Informer} reduces computational complexity while maintaining the ability to model long-range dependencies. Pretrained on large-scale datasets, it serves as a strong backbone for forecasting tasks. Inspired by vision Transformers, \textit{PatchTST} \cite{Nie2022} segments time series into non-overlapping patches and processes them using a hierarchical attention mechanism. This approach captures both local and global temporal patterns, making it well-suited for multivariate time series. \textit{PatchTST’s} ability to handle large-scale datasets with variable-length sequences exemplifies the flexibility of foundation models. \textit{ETSFormer} \cite{Woo2022} integrates traditional exponential smoothing methods with Transformer architectures, combining the strengths of statistical modeling and deep learning. By incorporating domain knowledge into the attention mechanism, \textit{ETSFormer} enhances interpretability and accuracy in forecasting applications. Drawing inspiration from the \textit{BERT} \cite{Devlin2018} model in NLP, \textit{TimeBERT} \cite{Wang2022} employs a masked time step prediction objective during pretraining. This enables the model to learn robust representations of temporal dependencies, which can be fine-tuned for diverse downstream tasks.

\subsubsection{Significance of Foundation Models for Time Series}

The rise of Transformer-based foundation models marks a paradigm shift in time series analysis. Unlike traditional approaches, where models are designed and trained for specific tasks or datasets, foundation models provide a unified framework for tackling diverse challenges. This shift offers several advantages:

\begin{enumerate}
    \item \textbf{Reduced Task-Specific Development Effort}: The pretraining-fine-tuning paradigm significantly reduces the need to design task-specific architectures from scratch.
    \item \textbf{Improved Performance on Low-Data Tasks}: By leveraging pretrained representations, foundation models achieve state-of-the-art performance even in scenarios with limited labeled data.
    \item \textbf{Cross-Domain Applicability}: Pretraining on heterogeneous datasets enables foundation models to transfer knowledge across domains, such as applying insights from financial datasets to healthcare or vice versa.
\end{enumerate}
The emergence of foundation models for time series represents a convergence of advances in machine learning and the growing demand for scalable, adaptable solutions to complex temporal problems. By leveraging the Transformer’s architectural strengths, these models are poised to become indispensable tools for researchers and practitioners across domains \cite{Wen2023}, \cite{Khan2021}, \cite{Lin2021}.

\section{Taxonomy}
\label{taxonomy}

\subsection{Challenges in Analyzing the Field}

The rapid proliferation of Transformer-based models for time series analysis has brought significant advancements, but it has also created challenges in organizing and understanding the field. Although some surveys exist \cite{Zhang2024}, \cite{Jiang2024}, \cite{Liang2024}, \cite{Miller2024}, \cite{Jin2023a}, they typically focus on broad overviews of time series forecasting or specific applications without delving deeply into the nuances of model architecture, training paradigms, or other critical taxonomic dimensions. This lack of depth makes it difficult for researchers and practitioners to gain a comprehensive understanding of the diverse landscape. Below, we outline the primary challenges in organizing and analyzing this rapidly evolving field.

\subsection{Lack of Detailed Taxonomy}

Time series Transformers are typically evaluated and categorized based on high-level criteria. However, a deeper taxonomy is required to understand the critical design choices underpinning these models. Key dimensions include:

\subsubsection{Model Architecture}

Pretrained models like \textit{GPT }\cite{OpenAI2023}\textit{, BERT }\cite{Devlin2018}, and their variants, originally developed for natural language processing (NLP), have been repurposed for time series analysis through fine-tuning or directly used without fine-tuning. Encoder-only, decoder-only, and hybrid encoder-decoder foundational models are trained from the ground up for time series tasks, with each architecture designed to capture different aspects of temporal dependencies and improve performance on tasks such as forecasting, anomaly detection, and classification. \textit{Tiny Time Mixers} are a lightweight, non-transformer model designed specifically for time series tasks, using efficient token and feature mixing to capture temporal dependencies \cite{Ekambaram2024}.

\subsubsection{Patch and Non-Patch Based Models}

Patch-based time series Transformers segment the input sequence into fixed-length \textit{“windows”} or \textit{“patches”}, which are treated as individual tokens \cite{Nie2022}. This approach allows the model to capture local temporal patterns within each segment before learning global dependencies across the entire sequence. Non-patch based time series Transformers process the entire time series as a continuous sequence without segmenting it into smaller units. These models leverage self-attention mechanisms to capture both short-term and long-term dependencies across the full sequence.

\subsubsection{Objective Functions}

Different objective functions can affect how well a model generalizes to unseen data. For example, while \textit{Mean Squared Error (MSE) }focuses on minimizing the prediction error for regression tasks \cite{Bao2007}, \textit{Negative Log-Likelihood (NLL)} provides probabilistic estimates that improve uncertainty modeling \cite{Christou2014}. Few surveys delve into how these objectives influence the choice of architecture and task suitability.

\subsubsection{Univariate vs. Multivariate Time Series Models}

Univariate models are designed for single-variable time series data, focusing on temporal dependencies within one sequence. They are commonly applied in tasks like univariate forecasting or anomaly detection. Multivariate models handle multiple time series or features at each time point, capturing complex relationships between variables and time. They are suitable for tasks like multivariate forecasting or cross-variable anomaly detection \cite{Bryan2021}, \cite{Schmidl2022}.

\subsubsection{Probabilistic vs. Non-Probabilistic Models}

Probabilistic models \cite{Syntetos2005} generate a distribution of possible outcomes, estimating uncertainty in their predictions. They are ideal for applications that require risk analysis or understanding prediction uncertainty. Non-Probabilistic models provide a single, fixed prediction for each time step \cite{Faes2020}.

\subsubsection{Model Scale and Complexity}

Lightweight models optimized for resource-constrained environments contrast sharply with large-scale foundation models capable of generalization across multiple domains \cite{Liu2024a}. The trade-offs between scalability and domain-specific performance are rarely explored in existing literature.

Without such a detailed taxonomy, researchers lack a clear framework for systematically comparing models, identifying gaps, and driving innovation. Rapid growth of Transformer-based models for time series analysis has led to significant advancements but has also created challenges in systematically organizing and understanding the field. While existing surveys \cite{Zhang2024}, \cite{Jiang2024}, \cite{Liang2024}, \cite{Miller2024}, \cite{Jin2023a} provide broad overviews or focus on specific applications, there remains a clear gap in detailed taxonomic frameworks that address the key architectural, training, and design choices critical for effective time series modeling. The lack of such a taxonomy complicates the comparison of models, hinders the identification of research gaps, and slows down innovation in this rapidly evolving area.

In the following sections of this paper, we aim to address these challenges by providing a structured and comprehensive taxonomy of time series Transformers. This includes a detailed examination of model architectures, patching strategies, objective functions, and distinctions between univariate and multivariate models, as well as the comparison of probabilistic vs. deterministic outputs. By exploring these dimensions, we aim to offer a clearer understanding of the field and provide a solid foundation for future research and application in time series analysis. High level overview the taxonomy is provided in the Figure \textit{8} and differences between different foundational models is provided in the Table \textit{2}.

 \begin{figure*}[t]  % Use figure* to span both columns
  \centering
  \includegraphics[width=\textwidth]{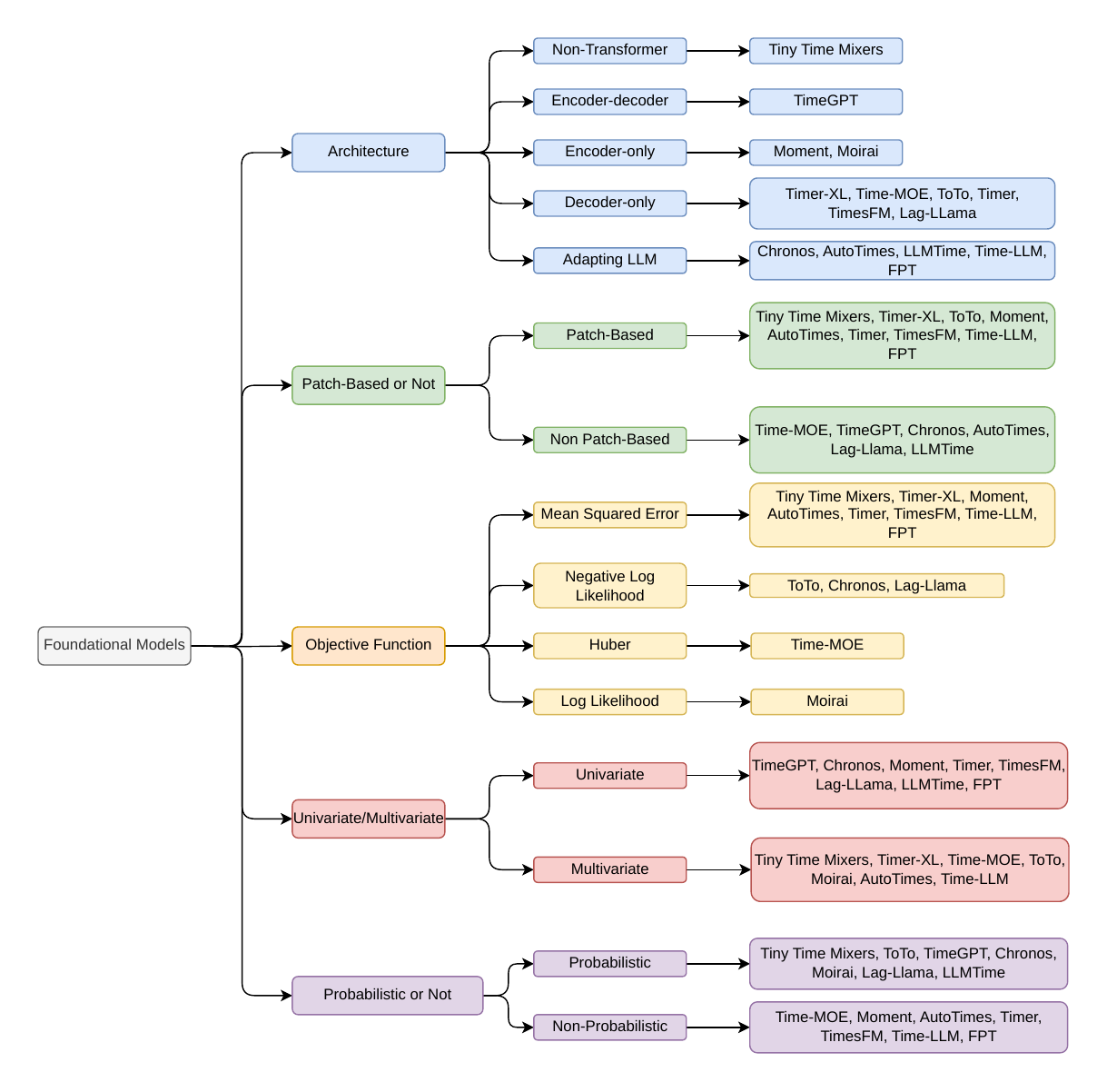}
  \caption{A Detailed Taxonomy: Exploring Model Architectures, Patch-Based vs. Non-Patch Approaches, Objective Functions, Univariate vs. Multivariate Models, Probabilistic vs. Non-Probabilistic Methods, and the Impact of Model Scale and Complexity on Performance}
  \Description{A Detailed Taxonomy: Exploring Model Architectures, Patch-Based vs. Non-Patch Approaches, Objective Functions, Univariate vs. Multivariate Models, Probabilistic vs. Non-Probabilistic Methods, and the Impact of Model Scale and Complexity on Performance}
  \label{fig:Fig8}
\end{figure*}

\section{Methodology}
\label{methodology}

\subsection{Model Architecture}

\subsubsection{Non-Transformer}

\textit{Tiny Time Mixers (TTM)} \cite{Ekambaram2024} is the only non-Transformer model discussed in this paper. It is based on a non-Transformer-based light-weight \textit{TSMixer }\cite{Ekambaram2023}, \cite{Chen2023} architecture and it incorporates innovations like adaptive patching to operate at varying patch lengths and number of patches, diverse resolution sampling to augment data to improve coverage across varying temporal resolutions, and resolution prefix tuning to handle pre-training on varied dataset resolutions with minimal model capacity. Additionally, it employs multi-level modeling to capture channel correlations and infuse exogenous signals during fine-tuning. \textit{TTM} supports channel correlations and exogenous signals, which are critical and practical business requirements in the context of multivariate forecasting, features lacking in many existing TS pretrained models.

\subsubsection{Encoder-decoder}

\textit{TimeGPT} \cite{Garza2023} is an encoder-decoder Transformer-based model designed for time-series forecasting, combining components from large language models (LLMs) with additional techniques like convolutional neural networks (CNNs) to enhance model performance. It incorporates positional encoding to capture the sequential nature of time-series data, multi-head attention to focus on different parts of the input simultaneously, and CNN layers for learning complex temporal patterns. The model also utilizes residual connections and layer normalization to stabilize training and improve convergence. \textit{TimeGPT} is trained on large, diverse time-series datasets and can be fine-tuned for specific forecasting tasks using zero-shot or few-shot learning methods. Its robust architecture makes it suitable for various applications, including load forecasting, web traffic prediction, financial forecasting, and healthcare data analysis, providing flexibility in handling complex, noisy time-series data across different domains. 

\subsubsection{Encoder-only}

The \textit{MOMENT} \cite{Goswami2024} architecture is a Transformer-based model designed for time series forecasting, addressing challenges posed by varied and smaller time series datasets. It is pre-trained on a diverse collection of datasets, referred to as the Time Series Pile, which includes data from domains such as electricity, traffic, and weather. The model uses a masked time series prediction task during pre-training, where patches of time series data are masked, and the model learns to predict the original values. The input time series are divided into fixed-length patches, which are embedded and processed through a Transformer encoder with modifications such as relative positional embeddings and instance normalization to handle varying time series characteristics. The architecture also includes a lightweight reconstruction head to map the transformed embeddings back to the original time series dimensions. Key features include handling variable-length time series, scalability through a simple encoder and minimal parameters, and flexibility in handling multivariate time series by processing each channel independently. \textit{MOMENT’s} design efficiently learns temporal representations from diverse time series data, making it suitable for forecasting, anomaly detection, and classification tasks across a wide range of domains.

The \textit{MOIRAI} \cite{Woo2024} model is a flexible and powerful approach for probabilistic multivariate time series forecasting, designed to handle data with varying frequencies and domains. It employs a masked encoder architecture with a patch-based input approach, which flattens multivariate time series into a single sequence and uses a masked token to facilitate autoregressive forecasting. The model outputs a mixture distribution, capturing predictive uncertainty, and incorporates improvements such as pre-normalization, \textit{RMSNorm} \textit{\cite{Zhang2019}}, query-key normalization \cite{Henry2020}, and \textit{SwiGLU} \cite{Shazeer2020} activation. To handle time series with different frequencies, \textit{MOIRAI} uses flexible patch sizes, applying larger patches for high-frequency data and smaller ones for low-frequency data. The Any-variate Attention mechanism allows it to process multivariate time series by flattening the data and using binary attention biases to maintain both permutation equivariance and invariance. For probabilistic forecasting, \textit{MOIRAI} leverages a mixture of distributions, including Student’s t-distribution, negative binomial, log-normal, and low-variance normal distributions, to model diverse uncertainties in time series data. The model is trained with a categorical cross-entropy loss function, treating the forecasting task as a regression via classification.\textit{ MOIRAI’s} ability to model complex uncertainty and handle varying time series frequencies and dimensions makes it a promising solution for diverse forecasting tasks.

\subsubsection{Decoder-only}

\textit{\textit{Timer-XL}} \cite{Liu2024b} is an advanced time series forecasting model that builds on generative decoder-based Transformer architectures. The model's key innovation lies in its \textit{TimeAttention} mechanism, which enables it to capture complex dependencies within and across time series while preserving temporal causality. \textit{Timer-XL} works by predicting the next \textit{“token”} (patch) in a time series, leveraging a decoder-only Transformer structure for univariate series. For multivariate forecasting, it extends this approach by defining tokens for each variable and learning dependencies between them, enhancing model performance over independent modeling of each channel.

A significant challenge in time series modeling is handling temporal order and variable relationships, which \textit{Timer-XL} addresses through position embeddings and a flexible attention mechanism. The model uses \textit{Rotary Position Embeddings (RoPE)} \cite{Su2024} for time dimension handling and learnable scalars for variable embedding, which ensures reflexivity and mitigates permutation invariance issues. The \textit{TimeAttention} mechanism incorporates both temporal and variable dependencies via a Kronecker product approach, ensuring the model attends to the appropriate information across time and variables. \textit{Timer-XL} is a versatile model capable of handling various forecasting scenarios, including cases with additional covariates, and offers a unified architecture for both univariate and multivariate tasks, making it a powerful tool for time series forecasting.

\textit{Time-MOE} \cite{Shi2024} is a mixture-of-experts-based, decoder-only Transformer model designed for time series forecasting, integrating several innovative components to enhance performance and scalability. It employs point-wise tokenization for efficient handling of variable-length sequences, with \textit{SwiGLU} \cite{Shazeer2020} gating to embed time series points. The model replaces the traditional feed-forward network with a \textit{Mixture-of-Experts (MoE)} layer, where a subset of expert networks is activated per input, reducing computation while maintaining performance. \textit{Time-MOE} also supports multi-resolution forecasting \cite{Zhang2024a}, allowing predictions at multiple time scales, which improves flexibility in handling different forecasting horizons. With its use of \textit{MoE} for computational efficiency and multi-resolution design for robust forecasting, \textit{Time-MOE} offers a scalable and flexible approach to time series forecasting.

\textit{Toto }\cite{Cohen2024} model is a decoder-only Transformer architecture designed for multivariate time series forecasting, incorporating key innovations to handle both time-wise (temporal) and space-wise (channel) dependencies. It utilizes a causal next-patch prediction task for efficient pre-training, leveraging techniques from large language models (LLMs) such as \textit{RMSNorm \cite{Zhang2019}}, pre-normalization, and \textit{SwiGLU }\cite{Shazeer2020} feed-forward layers to enhance learning capacity. A standout feature of \textit{Toto} is its \textit{Proportional Factorized Space-Time Attention} mechanism \cite{Bertasius2021}, which alternates between time-wise and space-wise attention blocks, balancing computational efficiency and accuracy. The model also introduces a probabilistic prediction head using a \textit{Student-T Mixture Model (SMM) }\cite{Peel2000}, enabling it to handle heavy-tailed distributions and outliers. \textit{Toto’s }ability to model complex distributions and quantify uncertainty through \textit{Monte Carlo} sampling makes it highly flexible and robust for time series forecasting across various domains.

The \textit{Timer} \cite{Liu2024c} architecture applies large language models (LLMs) to time series forecasting by leveraging the sequential nature of both time series data and language. It emphasizes the use of extensive, high-quality datasets, such as the \textit{Unified Time Series Dataset} \cite{Cyranka2023} (UTSD) with up to 1 billion time points across seven domains, and integrates \textit{LOTSA} \cite{Woo2024} for zero-shot forecasting. \textit{Timer} introduces a unified format called \textit{single-series sequence (S3)} to handle diverse time series data, allowing for easier preprocessing and normalization without the need for alignment across domains. The model is trained using generative pre-training in an autoregressive manner, predicting future time series values based on historical context. A decoder-only Transformer architecture is employed to maintain the sequential dependencies inherent in time series data, making it well-suited for time series forecasting tasks. This design ensures scalable, adaptable, and effective forecasting across varied datasets.

The \textit{TimesFM} \cite{Das2023} model is a Transformer-based architecture designed for efficient long-horizon time-series forecasting. It operates by breaking input time-series into non-overlapping patches, which reduces computational costs and enhances inference speed. The model uses a decoder-only architecture, where it predicts the next patch based on prior patches, enabling parallel prediction and avoiding inefficiencies in multi-step autoregressive decoding. A random masking strategy is employed during training to handle variable context lengths, allowing the model to generalize across different input sizes. The core of the model consists of stacked Transformer layers with causal masking to ensure that future time-steps do not influence past predictions. During inference, the model generates future time-series in an auto-regressive manner, where each forecast is concatenated with the input for further prediction. The loss function used is Mean Squared Error (MSE), optimizing for point forecasting accuracy. \textit{TimesFM’s} design offers flexibility in forecast horizons and context lengths, making it adaptable to various time-series datasets and suitable for zero-shot forecasting tasks.

\textit{Lag-LLaMA }\cite{Rasul2023} is a Transformer-based model for univariate probabilistic time series forecasting, built on the \textit{LLaMA} \cite{Touvron2023} architecture. It incorporates a specialized tokenization scheme that includes lagged features (past values at specified lags) and temporal covariates (e.g., day-of-week, hour-of-day), allowing it to handle varying frequencies of time series data. The model uses a decoder-only Transformer with causal masking and \textit{Rotary Positional Encoding (RoPE) }for sequential data processing. For forecasting, the output is passed through a distribution head that predicts the parameters of a probability distribution, providing not only point forecasts but also uncertainty quantification through probabilistic distributions. During training, it minimizes the negative log-likelihood of predicted distributions, and at inference, it generates multiple forecast trajectories through autoregressive decoding. 

\subsubsection{Adapting LLM}

\textit{Chronos }\cite{Ansari2024} is a framework that adapts language modeling techniques for probabilistic time series forecasting, utilizing a novel tokenization approach and a minimal modification of existing Transformer architectures. It converts continuous time series data into discrete tokens by first scaling the data using mean normalization and then quantizing it through a binning process, where values are assigned to predefined bins. The model uses a categorical cross-entropy loss function to train on these tokenized sequences, allowing it to learn multimodal distributions. The core architecture of \textit{Chronos} is based on the \textit{T5} \cite{Raffel2020} encoder-decoder model, though it can also be adapted to decoder-only models like \textit{GPT-2} \cite{Radford2019}. This architecture remains largely unchanged from standard language models, with adjustments made only to the vocabulary size to account for the quantization bins. During forecasting, \textit{Chronos} generates probabilistic predictions by autoregressively sampling from the predicted distribution, enabling the generation of prediction intervals and uncertainty quantification. Its flexibility in handling diverse time series datasets, ability to model arbitrary distributions, and minimal changes required to implement existing language models make \textit{Chronos} a scalable and efficient solution for time series forecasting.

\textit{AutoTimes} \cite{Liu2024d} is an approach that adapts large language models (LLMs) for multivariate time series forecasting by treating time series as sequences of tokens. The model processes time series data by splitting it into non-overlapping segments, where each segment represents a single variate and is treated independently. Temporal context is provided through timestamp position embeddings, which align the segments based on their respective time points. \textit{AutoTimes} uses an autoregressive approach, predicting the next time series segment iteratively, conditioning each prediction on previous values. The model can handle varying lookback and forecast lengths with a single unified model, avoiding the need for multiple models for different forecasting horizons. The training objective is based on minimizing mean squared error (MSE) between predicted and actual segments. Key innovations include segment-wise tokenization, timestamp embeddings for temporal context, and autoregressive multi-step forecasting. Overall, \textit{AutoTimes} leverages the flexibility and scalability of LLMs to efficiently forecast multivariate time series data with varying time horizons.

\textit{LLMTime }\cite{Gruver2023} is a novel architecture for time series forecasting that frames the problem as next-token prediction, similar to text generation in language models. In this model, time series data is encoded as a string of numerical digits, where each time step is represented by its individual digits separated by spaces. The model performs forecasting by tokenizing the time series in this manner, enabling it to treat time series forecasting as a sequence prediction task, akin to predicting the next word in a sentence.

\textit{TIME-LLM} \cite{Jin2024} is a reprogramming framework that adapts large language models (LLMs) for time series forecasting without fine-tuning the backbone model. It operates by transforming time series data into text prototype representations, making it compatible with the LLM’s natural language processing capabilities. The input time series are processed into univariate sequences, normalized, patched, and embedded before being reprogrammed with learned text prototypes. To enhance the model’s reasoning, the input is augmented with domain-specific prompts. The transformed data is then projected through the LLM to generate forecasts. The framework utilizes a frozen LLM, with only the input transformation and output projection parameters updated, allowing for efficient few-shot and zero-shot learning while outperforming specialized forecasting models. This approach maintains minimal resource requirements and does not necessitate extensive fine-tuning, making it highly efficient and versatile for time series forecasting across various tasks.

The \textit{Frozen Pretrained Transformer (FPT)} \cite{Lu2022} model leverages pre-trained language or vision models, such as \textit{GPT} \cite{OpenAI2023}, \textit{BERT }\cite{Devlin2018}, and \textit{BEiT }\cite{Bao2022}, for time series analysis by freezing the self-attention and feedforward layers of the Transformer architecture while fine-tuning the positional embedding, layer normalization, and output layers. The model uses a redesigned input embedding layer to project time series data into the required dimensions, employing linear probing to reduce training parameters. Additionally, a reverse instance normalization block is added to further normalize the data and facilitate knowledge transfer. To enhance the model's ability to capture local semantic information, patching is applied by aggregating adjacent time steps into single patch-based tokens, allowing for extended historical time horizons without increasing token length. This architecture enables effective cross-domain knowledge transfer from language and image models to time series forecasting tasks.

\subsection{Patching vs. Non-Patching Approaches}

\subsubsection{Patch-based}

\textit{Tiny Time Mixers }\cite{Ekambaram2024} use non overlapping windows as patches during pre-training phase. \textit{Timer-XL} \cite{Liu2024b} uses patch-level generation based on long-context sequences for multivariate forecasting tasks. \textit{Toto} \cite{Cohen2024} is pre-trained on the next-patch prediction task. In the \textit{MOMENT} \cite{Goswami2024} architecture, patching involves dividing time series data into fixed-length segments, embedding each segment, and using a masked time series prediction task to train the model. \textit{MOIRAI} \cite{Cyranka2023} follows a (non-overlapping) patch-based approach to modeling time series with a masked encoder architecture. In \textit{AutoTimes }\cite{Liu2024d}, time series data is split into non-overlapping segments, with each segment representing a single variate, which are then treated as individual tokens. This tokenization process helps the model capture inter-variate correlations while simplifying the temporal structure for the LLM. \textit{Timer} \cite{Liu2024c} uses a segment-based approach, where time series data is processed as \textit{single-series sequences }(S3), treating each time series as a sequence of tokens. This method enables the model to handle diverse time series without requiring time alignment across different datasets. \textit{TimesFM} \cite{Das2023} uses a patching approach, where the input time-series is split into non-overlapping patches, similar to tokens in NLP Transformers. This reduces computational costs and enhances model efficiency by processing multiple time steps in parallel. In the \textit{TIME-LLM} \cite{Jin2024} model, patching is used to process time series data by dividing multivariate time series into univariate patches, which are then reprogrammed with learned text prototypes. This approach helps align the input time series with the language model’s capabilities for efficient forecasting. The \textit{Frozen Pretrained Transformer (FPT)} \cite{Lu2022} model uses patching to extract local semantic information by aggregating adjacent time steps into single patch-based tokens, allowing for a larger input historical time horizon while maintaining token length.

\subsubsection{Non Patch-based}

\textit{\textit{Time-MOE}} \cite{Shi2024} uses point-wise tokenization for time series embedding. \textit{TimeGPT}  \cite{Garza2023} does not use patch technique. In \textit{Chronos} \cite{Ansari2024}, the tokenization process is different from the patch technique, as it involves discretizing the time series values into bins rather than splitting the data into fixed-size patches.\textit{ Lag-LLaMA}  \cite{Rasul2023} does not use patching or segmentation but tokenizes time series data by incorporating lagged features and temporal covariates. Each token consists of past values at specified lag indices and additional time-based features, making it capable of handling varying time series frequencies. \textit{LLMTime} \cite{Gruver2023} does not use patching or segments; instead, it encodes the time series as a string of numerical digits, treating each time step as a sequence of tokens.

\subsection{Objective Function}

\subsubsection{Mean Squared Error}

\textit{Tiny Time Mixers} \cite{Ekambaram2024} are pre-trained with mean squared error (MSE) loss calculated over the forecast horizon. \textit{Timer-XL} \cite{Liu2024b} are pre-trained with mean squared error (MSE) loss. \textit{MOMENT} \cite{Goswami2024} uses mean squared error (MSE) as the objective function. The objective function in \textit{AutoTimes} \cite{Liu2024d} is based on minimizing the mean squared error (MSE) between the predicted and actual time series segments.

The objective function in \textit{Timer} \cite{Liu2024c} is based on autoregressive pre-training, where the model is trained to predict the next time series value given the previous ones. The loss function typically minimizes the difference between the predicted and actual values, often using mean squared error (MSE) to optimize model parameters. The objective function of the \textit{TimesFM} \cite{Das2023} model is based on Mean Squared Error (MSE), where the model is trained to minimize the average MSE between the predicted output patches and the actual future values. The objective function in the \textit{TIME-LLM} \cite{Jin2024} model minimizes the mean squared error (MSE) between the predicted time series values and the ground truth over the forecast horizon. The model aims to optimize the accuracy of future time step predictions by projecting the language model’s output to match the target values. The \textit{Frozen Pretrained Transformer (FPT)} \cite{Lu2022} model's objective function focuses on fine-tuning the pre-trained Transformer parameters for time series tasks, optimizing the model's performance through tasks such as forecasting. The model is trained to minimize the loss between the predicted and actual time series values, typically using mean squared error (MSE).

\subsubsection{Huber Loss}

\textit{\textit{Time-MOE}} \cite{Shi2024} uses Huber and Auxiliary objective functions. Huber Loss \cite{Meyer2021} is used to improve robustness to outliers and ensure stability during training. Auxiliary Loss is used to prevent expert imbalances (routing collapse) and ensure that the mixture-of-experts layer is used effectively, an auxiliary loss is added that balances expert usage. The final loss combines auto-regressive forecasting errors from multi-resolution projections and the auxiliary balance loss.

\subsubsection{Negative Log Likelihood}

\textit{Toto} \cite{Cohen2024} minimizes the negative log-likelihood of the next predicted patch with respect to the distribution output of the model. \textit{Chronos} \cite{Ansari2024} is trained to minimize the cross entropy between the distribution of the quantized ground truth label and the predicted distribution. The objective function in the \textit{Lag-LLaMA} \cite{Rasul2023}model minimizes the negative log-likelihood of the predicted distribution over future time steps. This approach enables the model to forecast probabilistic outcomes by predicting the parameters of a chosen distribution for each time step.

\subsubsection{Log Likelihood}

The objective in \textit{MOIRAI} \cite{Cyranka2023} model is to forecast the predictive distribution by predicting the distribution parameters through a learned model that takes the past time series and covariate data as input and maximizes the log-likelihood. The objective function in \textit{LLMTime} \cite{Gruver2023}is based on maximizing the likelihood of the next token in the time series sequence, treating time series forecasting as a next-token prediction task. During training, the model minimizes the negative log-likelihood of the predicted token distributions over the actual future time steps. \textit{LLMTime} \cite{Gruver2023} is not pre-trained, so there is no objective function.

\subsection{Univariate vs. Multivariate}

\subsubsection{Univariate}

\textit{TimeGPT} \cite{Garza2023} can only handle univariate data out of the box. \textit{Chronos} \cite{Cohen2024} can only handle univariate data out of the box. \textit{MOMENT} \cite{Goswami2024} can only handle univariate data out of the box. \textit{MOIRAI} \cite{Cyranka2023} can also handle both univariate and multivariate time series data.\textit{ Timer} \cite{Liu2024c} primarily supports univariate time series by normalizing and merging diverse time series into a unified pool of univariate sequences. The model does not explicitly mention support for multivariate time series, as it focuses on handling a wide variety of univariate datasets across multiple domains. \textit{TimesFM} \cite{Das2023} appears to focus on univariate time series forecasting, with no explicit support for multivariate data. While the model's architecture could theoretically accommodate multivariate time series, this adaptation is not explicitly outlined in the design. The \textit{Lag-LLaMA} \cite{Rasul2023} model is designed specifically for univariate time series forecasting. It does not natively support multivariate time series, as it focuses on forecasting a single time series with probabilistic outputs. The objective function in \textit{LLMTime} \cite{Jin2024} is based on maximizing the likelihood of the next token in the time series sequence, treating time series forecasting as a next-token prediction task. During training, the model minimizes the negative log-likelihood of the predicted token distributions over the actual future time steps. The \textit{Frozen Pretrained Transformer (FPT)} \cite{Lu2022} model supports both univariate and multivariate time series forecasting. It processes multivariate time series by treating each variable independently after transforming the input, allowing it to handle multiple time series variables simultaneously. This is same as univariate forecasting.

\subsubsection{Multivariate}

\textit{\textit{Tiny Time Mixers}} \cite{Ekambaram2024} are pre-trained with mean squared error (MSE) loss calculated over the forecast horizon. \textit{Timer-XL} \cite{Liu2024b} can handle both univariate and multivariate time series data as well as covariate-informed forecasting. \textit{Time-MOE} \cite{Shi2024} can handle both univariate and multivariate time series data. \textit{Toto} \cite{Cohen2024} can also handle both univariate and multivariate time series data. In \textit{AutoTimes }\cite{Liu2024d}, multivariate time series are handled by processing each variate (or column) independently, meaning it primarily supports univariate time series forecasting. Each variate is treated as a separate sequence to capture individual time series patterns, but inter-variate dependencies are not explicitly modeled within the framework. Probabilistic vs. Non-Probabilistic Models

\subsection{Probabilistic vs. Non-Probabilistic Models}

\subsubsection{Probabilistic}

\textit{Toto} \cite{Cohen2024}, \textit{Chronos} \cite{Ansari2024} and \textit{TimeGPT} \cite{Garza2023} can generate probabilistic forecasts. \textit{MOIRAI} \cite{Cyranka2023} can also generate probabilistic forecasts. \textit{Lag-LLaMA} \cite{Rasul2023} model also supports probabilistic forecasts. It predicts the parameters of a probability distribution for the next time step, allowing for uncertainty quantification in its predictions.\textit{ LLMTime} \cite{Jin2024} supports probabilistic forecasts by generating multiple samples and using statistical methods such as median or quantiles to form point estimates or uncertainty intervals. It models time series data as continuous distributions, enabling high-resolution probabilistic forecasting.

\subsubsection{Non-Probabilistic}

\textit{Tiny Time Mixers} \cite{Ekambaram2024} only generates point forecasts and doesn’t facilitate probabilistic forecasts as it does not use distribution head in its architecture. \textit{Timer-XL} \cite{Liu2024b} only generates point forecasts. \textit{Time-MoE} \cite{Shi2024} can only generate point forecasts. \textit{MOMENT} \cite{Goswami2024} cannot generate probabilistic forecasts. \textit{Autotimes} \cite{Liu2024d} does not support probabilistic forecasts. \textit{Timer} \cite{Liu2024c} does not explicitly support probabilistic forecasting. The model is designed to predict specific future points rather than providing a distribution over possible outcome. The \textit{TimesFM} \cite{Das2023} model does not natively support probabilistic forecasting, as it primarily focuses on point forecasting using a mean squared error (MSE) loss function. The \textit{TIME-LLM} \cite{Jin2024} model supports multivariate time series forecasting by partitioning the input into multiple univariate series, processing them independently. Each univariate series is transformed and reprogrammed into the language model for forecasting, enabling multivariate support without altering the backbone LLM. The \textit{Frozen Pretrained Transformer (FPT)} \cite{Lu2022} model does not explicitly focus on probabilistic forecasting. It primarily generates point estimates for time series forecasting, with no mention of uncertainty quantification or probabilistic prediction in the model's architecture.

\subsection{Model Scale and Complexity}

\textit{Tiny Time Mixers} \cite{Ekambaram2024} is of 1 million parameter model which is pre-trained on 1 billion datapoints. \textit{Timer-XL} \cite{Liu2024b} is pre-trained on 1 billion diverse data points. \textit{Time-MOE} model \cite{Shi2024}has 2.4 billion parameters and it pre-trained on 309 billion data points. \textit{Toto} \cite{Cohen2024} is a 103 million parameter model and it pre-trained on 1 trillion data points. \textit{TimeGPT} model \cite{Garza2023} parameter number is not disclosed in the paper and the model was pre-trained on 100 billion data points. Several \textit{Chronos} \cite{Ansari2024} models of different sizes (8, 46, 201, 710 million parameters) are trained on 84 billion data points. Several \textit{MOMENT} \cite{Goswami2024} models of different sizes (40, 125, 385 million parameters) are pre-trained on 1.13 billion data points. Several \textit{MOIRAI} models \cite{Woo2024} of different sizes (14, 91, 311 million parameters) are pre-trained on 27 billion data points. Different foundational LLM models (\textit{GPT-2} \cite{Radford2019}124M, \textit{OPT}-350M, 1.3B, 2.7B, 6.7B \cite{Zhang2022}, LLaMA-7B \cite{Touvron2023}) are pre-trained on diverse datasets for \textit{Autotimes} model\cite{Liu2024d}. Several \textit{Timer} models \cite{Liu2024c} of different sizes (29, 50, 67 million parameters) are pre-trained on 28 billion data points. \textit{TimesFM} \cite{Das2023} is a 200 million parameter model which is pre-trained on 200 billion data points. \textit{Lag-Llama} \cite{Rasul2023}is a 200 million parameter model which is pre-trained on 352 million data points. \textit{LLMTime} \cite{Gruver2023} is a transformer-based model that leverages large-scale pre-training on diverse time series datasets, with models such as \textit{LLaMA-1} 70B \cite{Touvron2023}, \textit{GPT-3 }\cite{Brown2020}, and \textit{GPT-4 }\cite{OpenAI2023}. Unlike methods that leverage LLM backbones, \textit{LLMTime} method is entirely zero-shot and does not require finetuning. \textit{TIME-LLM} model \cite{Jin2024} uses large pre-trained language models such as \textit{LLaMA} \cite{Touvron2023} and \textit{GPT-2 }\cite{Radford2019} as the backbone. These models are reprogrammed for time series forecasting without fine-tuning the core parameters, leveraging their ability to process sequences and generate forecasts. \textit{Frozen Pretrained Transformer (FPT) }model \cite{Lu2022} uses \textit{GPT-2: 1.5B }\cite{Radford2019}, \textit{BERT} \cite{Bao2022}and \textit{BEiT} models \cite{Bao2022} are underlying models.

\begin{table*}[t]
\caption{Comparison of Time Series Models: Architectures, Parameters, and Characteristics}
\label{tab:models-comparison}
\resizebox{\textwidth}{!}{%
\footnotesize
\begin{tabular}{|l|l|l|l|l|l|l|l|}
\hline
\textbf{Model} & \textbf{Architecture} & \textbf{Univariate/Multivariate} & \textbf{Parameters} & \textbf{Data} & \textbf{Patch} & \textbf{Loss} & \textbf{Probabilistic} \\
\hline
Tiny Time Mixers & Non-Transformer & Both & 1M & 1B & Yes & MSE & No \\
\hline
Timer-XL & Decoder & Both & Unknown & 1B & Yes & MSE & No \\
\hline
Time-MoE & Decoder & Both & 2.4B & 309B & No & Huber & No \\
\hline
Toto & Decoder & Both & 103M & 1T & Yes & Neg Log-likelihood & Yes \\
\hline
TimeGPT & Encoder-Decoder & Univariate & Unknown & 100B & No & Unknown & No \\
\hline
Chronos & Encoder-Decoder & Univariate & 8M, 46M, 201M, 710M & 84B & No & Cross Entropy & Yes \\
\hline
MOMENT & Encoder & Univariate & 40M, 125M, 385M & 1.13B & Yes & MSE & No \\
\hline
MOIRAI & Encoder & Both & 14M, 91M, 311M & 27B & Yes & Log-likelihood & Yes \\
\hline
AutoTimes & LLM & Both & GPT-2, OPT-350M, LLaMA-7B & Same as LLM & Yes & MSE & No \\
\hline
Timer & Decoder & Univariate & 29M, 50M, 67M & 28B & Yes & MSE & No \\
\hline
TimesFM & Decoder & Univariate & 200M & 100B & Yes & MSE & No \\
\hline
Lag-Llama & Decoder & Univariate & 200M & 352M & No & Neg Log-likelihood & Yes \\
\hline
LLMTime & LLM & Univariate & Llama 70B, GPT-3/4 & Same as LLM & No & Unknown & Yes \\
\hline
Time-LLM & LLM & Both & Llama-1 70B, GPT-3 & Same as LLM & Yes & MSE & No \\
\hline
FPT & LLM & Univariate & GPT-2: BERT, BEiT & Same as LLM & Yes & MSE & No \\
\hline
\end{tabular}%
}
\end{table*}

\section{Issues} \label{issues}

\textit{Tiny Time Mixers} \textit{(TTM) }\cite{Ekambaram2024} is currently focused solely on forecasting tasks and do not support capabilities across multiple downstream tasks, including classification, regression, and anomaly detection right out of the box. Another limitation of \textit{TTM} is the need to train different models for different context length settings. Due to its non-Transformer-based architecture, \textit{TTM} is sensitive to context lengths. \textit{Tiny Time Mixers} only generates point forecasts and doesn’t facilitate probabilistic forecasts as it does not use distribution head in its architecture.

\textit{Timer-XL} \cite{Liu2024b} needs to improve the context utilization and computational efficiency. \textit{Timer-XL} necessitates iterative generation for long-term forecasting, which may lead to error accumulation and inflexibility in the output length. \textit{Timer-XL} doesn’t incorporate multi-resolution patches for input and output series. Data used for pre-training \textit{Toto} model \cite{Cohen2024} was internally generated by Datadog.

The application of \textit{TimeGPT} \cite{Garza2023} in load forecasting reveals several limitations, particularly when dealing with scarce historical data. While the model shows promise, it requires fine-tuning to achieve effective performance, as it cannot be used directly without adaptation. Its reliance solely on historical load data restricts its ability to incorporate external factors, such as Numerical Weather Prediction (NWP) data, which could improve forecasting accuracy. Furthermore, despite being trained on a vast and diverse set of datasets, the mismatch between these datasets and the specific characteristics of load data leads to reduced performance. While \textit{TimeGPT} excels in short-term forecasting with limited data, it underperforms when rich historical load data is available, where traditional machine learning models tend to outperform it. Additionally, the model's performance can vary depending on the similarity between the training and target datasets, necessitating careful evaluation through validation sets to determine its suitability for specific forecasting tasks.

The computational expense of handling long sequences and large datasets may also present scalability challenges, especially when tokenization and autoregressive sampling become resource-intensive in \textit{Chronos }\cite{Ansari2024}.

The \textit{MOMENT} model \cite{Goswami2024}, while effective for time series forecasting, faces challenges in high-stakes applications like healthcare. Its reliance on diverse and high-quality training data means biased data can lead to skewed predictions. The model's normalization may obscure key differences in vertically shifted time series, limiting its performance. Additionally, a lack of interpretability and the need for domain-specific fine-tuning pose barriers to trust and decision-making. Ethical concerns regarding equitable access to \textit{MOMENT} also arise, as unequal access could worsen healthcare disparities. Addressing these issues through fine-tuning, improving explainability, and ensuring fair access is crucial for responsible use.

Despite its strong performance in both in-distribution and out-of-distribution forecasting, the \textit{MOIRAI} \cite{Woo2024} model has several limitations that need attention. First, the model's hyperparameter tuning was limited due to resource constraints, and more efficient techniques, could potentially improve its results. The multi-patch size mapping used for handling cross-frequency learning is somewhat heuristic, indicating that a more flexible and elegant approach could be developed in future work. Additionally, the architecture currently lacks scalability for high-dimensional time series, and improving methods for extending Transformer input lengths could help address this issue. The model’s masked encoder structure also presents an opportunity to explore latent diffusion architectures as a potential enhancement. In terms of data, while the LOTSA dataset used for training is diverse, incorporating even more variation across domains and frequencies could improve model performance. Lastly, the future integration of multi-modal inputs, such as tabular or textual data, would further unlock the potential of \textit{MOIRAI} for universal time series forecasting tasks.

\textit{AutoTimes} \cite{Liu2024d} faces several limitations, including its lack of support for probabilistic forecasting, as it maps time series segments to latent embeddings instead of discrete tokens, limiting its ability to capture the uncertainty inherent in time series data. Additionally, while the model leverages autoregressive methods for forecasting, it has not yet fully explored advanced embedding and projection layers for more compatible time series tokenization. The approach also lacks flexibility for handling real-world multimodal time series datasets, such as those combining news and stock data or logs and measurements. Future work aims to improve the model's adaptability and explore low-rank adaptation techniques and more sophisticated embedding layers to better handle complex token transitions and multimodal data.

Despite its strong performance in forecasting, imputation, and anomaly detection, the \textit{Timer} model \cite{Liu2024c} has several limitations that need to be addressed. Key challenges include improving its ability to generalize in zero-shot scenarios, as well as enhancing its support for probabilistic forecasting and long-context predictions. While Timer excels in autoregressive generation and utilizes a large, diverse time series dataset for pre-training, its current architecture does not accommodate probabilistic outputs, limiting its flexibility in uncertain environments. Additionally, the model's performance in long-term forecasting and its ability to adapt to a wider range of real-world applications could benefit from further refinement. These areas point to crucial directions for future research and model development.

The \textit{TimesFM} model \cite{Liu2024b}, while effective for time series forecasting, faces limitations in terms of its support for multivariate data, as it primarily caters to univariate time series without explicit adaptation for handling multiple variables. Additionally, its reliance on fixed-length patches introduces trade-offs between computational efficiency and flexibility, particularly when dealing with shorter time series. The model also employs an autoregressive decoding process during inference, which, while effective for generating forecasts, may become inefficient when handling varying forecast horizons. Furthermore, while \textit{TimesFM} demonstrates flexibility in its output patch length and context handling, it does not explicitly support probabilistic forecasting, limiting its ability to capture and convey forecasting uncertainty. Lastly, there is a need for further refinement in how it handles longer context lengths, especially for applications requiring long-term predictions across diverse datasets.

While \textit{Lag-LLaMA} \cite{Rasul2023} demonstrates impressive performance in univariate probabilistic time series forecasting, particularly in zero-shot generalization and few-shot adaptation, its primary limitation lies in its lack of support for multivariate time series forecasting. Additionally, while it achieves strong results across diverse datasets, the model's scalability remains a challenge, and further work is needed to expand its capabilities, such as scaling up the model size and addressing the limitations of current time series datasets, which are still too small for optimal training. Future research should focus on developing models that can capture the complex dynamics of multivariate datasets and improve model performance by leveraging larger, more diverse training corpora.

\textit{LLMTime }\cite{Gruver2023}, leveraging LLMs for time series forecasting, benefits from pretraining on vast datasets, enabling zero-shot forecasting with minimal computational resources. While the model excels at generalizing patterns and integrating various capabilities such as question answering, it faces challenges like limited context windows, especially for multivariate time series, which are less suited for smaller contexts. Advances in expanding context windows to tens of thousands of tokens offer promising solutions, though arithmetic and recursive operations, often weak in LLMs, could restrict performance in certain complex time-series scenarios. Nonetheless, \textit{LLMTime’s} ability to frame forecasting as natural language generation opens up new avenues for fine-tuning and further model enhancement, making it an exciting direction for future research in time series prediction.

\textit{TIME-LLM} \cite{Jin2024} demonstrates the potential of leveraging pre-trained large language models (LLMs) for time series forecasting by reprogramming time series data into text-like representations and using natural language prompts to enhance the model’s reasoning. While the model shows impressive performance, surpassing expert models in some cases, there remains a need to further refine the reprogramming methods and enrich LLMs with domain-specific knowledge, such as through continued pre-training on time series data. The framework opens possibilities for multimodal models capable of joint reasoning across time series, natural language, and other data types. Future work should also focus on broadening LLMs’ capabilities, not just for time series, but across a variety of other complex tasks.

The \textit{Frozen Pretrained Transformer (FPT) }model \cite{Lu2022} demonstrates strong performance in time series analysis by leveraging pre-trained models from NLP and Computer Vision (CV), providing a unified framework that facilitates training for a range of tasks. While it performs competitively with state-of-the-art methods, it still lags behind models like \textit{N-BEATs }\cite{Oreshkin2020} in zero-shot scenarios across some datasets. Additionally, the exploration of the universality of Transformers, specifically the connection between self-attention and Principal Component Analysis (PCA), is still in its early stages. The model's performance could be further improved through parameter-efficient fine-tuning techniques and a deeper understanding of transformers through n-gram modeling as per the paper. These directions highlight the potential for enhancing both the adaptability and theoretical understanding of Transformers in time series forecasting.

\section{Conclusion} \label{conclusion}

% In conclusion, this paper presents a comprehensive exploration of the evolving landscape of time series Transformers, emphasizing the critical design choices that influence their performance and applicability across a variety of tasks. By offering a detailed taxonomy that encompasses model architecture, patch-based and non-patch approaches, objective functions, univariate versus multivariate models, and the distinction between probabilistic and non-probabilistic methods, we provide a clear framework for understanding the complexities and trade-offs involved in selecting and applying these models. Moreover, we highlight the ongoing challenges and opportunities in balancing model scale, complexity, and domain-specific performance. As time series Transformers continue to advance, a deeper understanding of these key dimensions will be crucial for driving innovation and improving the efficiency and effectiveness of time series analysis in real-world applications. Future research should aim to refine these frameworks, addressing gaps in scalability, generalization, and task-specific adaptation to ensure that these powerful models can meet the diverse needs of industries ranging from healthcare to finance and beyond.

This paper offers a thorough and methodical investigation into the rapidly advancing domain of time series Transformers, illuminating the critical design choices that shape their performance, efficiency, and applicability across a diverse array of tasks. As deep learning continues to transform the landscape of time series forecasting and analysis, gaining a nuanced understanding of the trade-offs inherent in Transformer-based architectures has become indispensable for both researchers and industry practitioners. This understanding is vital for harnessing the full potential of these models in real-world scenarios. 
To foster this understanding, we have developed a comprehensive taxonomy that systematically categorizes and examines the core elements of time series Transformers. This taxonomy addresses essential dimensions such as architectural variations, the differentiation between patch-based and non-patch-based approaches, the choice of objective functions, the distinctions between univariate and multivariate modeling frameworks, and the contrast between probabilistic and non-probabilistic methods. By establishing this structured framework, we aim to clarify the complexities of designing and implementing these models, providing actionable insights that can guide their selection, optimization, and deployment for specific use cases. This classification not only simplifies the process of model selection but also helps identify the strengths and weaknesses of various approaches, empowering practitioners to make more informed decisions when applying Transformers to time series problems.

In addition to the taxonomy, we have emphasized the significant challenges and promising opportunities that are shaping the future of time series Transformers. Among the most pressing challenges is the trade-off between model scale and computational efficiency. While larger Transformers often deliver superior predictive accuracy, their substantial computational requirements can render them impractical for environments with limited resources. Consequently, optimizing these models to achieve a balance between performance and computational feasibility remains a critical area of research. Other persistent challenges include addressing long-term dependencies, improving model interpretability, and enhancing data efficiency. These issues call for continued innovation in areas such as hybrid architectures, advanced attention mechanisms, and self-supervised learning techniques.

As the field evolves, achieving scalability, ensuring generalization across diverse datasets, and enabling adaptability to domain-specific tasks will be paramount to unlocking the full potential of time series Transformers in real-world applications. Future research efforts should concentrate on refining existing frameworks by developing methods that enhance model robustness, mitigate overfitting, and improve cross-domain transferability. Incorporating domain-specific knowledge, integrating physics-informed models, and leveraging reinforcement learning strategies could further enhance the reliability, interpretability, and performance of these models.
Ultimately, this work seeks to contribute to the ongoing progress in time series analysis by providing researchers, engineers, and industry professionals with a deeper understanding of time series Transformers and their practical applications. By addressing current challenges and seizing emerging opportunities, the research community can pave the way for the development of next-generation models that are more scalable, interpretable, and impactful. These advancements will enable the resolution of complex real-world problems, fostering innovation and improving outcomes across a wide range of industries.

\section{Acknowledgments}

We would like to acknowledge OpenAI ChatGPT-4o for assisting in the correction of grammar and other details in this paper. Tool is used to improve the clarity and readability of our manuscript. Additionally, we acknowledge the use of icons from Flaticon (www.flaticon.com) for creating some of the figures in this paper.

% \bibliographystyle{ACM-Reference-Format}
% \bibliography{sample-base}  % Replace 'your-bib-file' with your .bib file's name (without the .bib extension)

\bibliographystyle{ACM-Reference-Format}

% Insert your .bbl content here directly:

\end{document}